\documentclass{article}

\usepackage[preprint]{neurips_2026}

\usepackage[utf8]{inputenc}
\usepackage[T1]{fontenc}
\usepackage{hyperref}
\usepackage{url}
\usepackage{booktabs}
\usepackage{amsfonts}
\usepackage{amsmath}
\usepackage{amssymb}
\usepackage{nicefrac}
\usepackage{microtype}
\usepackage{xcolor}
\usepackage{graphicx}
\usepackage{subcaption}
\usepackage{multirow}
\usepackage{algorithm}
\usepackage{algorithmic}
\usepackage{wrapfig}
\usepackage{enumitem}
\usepackage{tikz}
\graphicspath{{images/}}
\usepackage[table]{xcolor}

\newcommand{\ours}{\textsc{PRISM}}

\title{Compositional Reward Models for Conditional Medical Image Generation}

\author{%
  Aayush Kumar Tyagi \\
  Yardi School of Artificial Intelligence \\
  India Institute Of Technology, Delhi\\
  \texttt{aiz218615@scai.iitd.ac.in} \\
  \And
  Prathosh A.P. \\
  Department of Electrical Communication Engineering \\
  Indian Institute of Science, Bengaluru \\
  \texttt{prathosh@iisc.ac.in} \\
  \AND
  Mausam \\
  Department of Computer Science and Engineering\\
  Indian Institute of Technology, Delhi \\
  \texttt{mausam@cse.iitd.ac.in}
}

\begin{document}

\maketitle

\begin{abstract}


Acquiring high quality annotated medical image data is critical for training deep learning models; however, annotation is expensive, time consuming, and often requires domain expertise. Conditional diffusion models, such as ControlNet, offer a promising alternative by generating images conditioned on semantic masks and text. However, existing approaches frequently fail to capture fine grained properties (e.g., intensity and texture), as well as higher level semantic consistency expected by domain experts, limiting their effectiveness for downstream tasks. Recent attempts to address these issues using reinforcement learning fine-tuning remain limited due to the reliance on a single scalar reward, which conflates diverse failure modes and provides weak corrective signals.

We propose \ours{}, a Compositional Reward Model (CRM) framework for conditional medical image generation. Instead of assigning a single reward, we decompose image quality into verifier grounded stages, each evaluating a distinct aspect of correctness from fine to coarse properties, including low level attributes (intensity and texture), structural alignment with conditioning inputs, and high level semantic fidelity. These stage wise rewards are composed through a Hierarchical Constrained Propagation (HCP) mechanism that enforces a fine to coarse notion of correctness, ensuring that lower level deficiencies are resolved before higher level rewards are accrued, preventing easier objectives from masking critical failures.

We evaluate \ours{} across three datasets spanning diverse medical imaging tasks: PanNuke (multi-class cell segmentation), CeDeM (villi/crypt detection and measurement), and ISIC (skin lesion classification). Training downstream models with data generated by \ours{} yields improvements over closest baselines, including a $2.3\%$ increase in mDice on PanNuke, a $8.5\%$ reduction in Mean Relative Error (MRE) on CeDeM, and increases ISIC F1 by $5.9\%$.

\end{abstract}
\section{Introduction}
\label{sec:introduction}

Deep learning models for medical imaging rely heavily on large scale, finely annotated datasets, often requiring expert pathologists for pixel level supervision. Acquiring such data is both expensive and time consuming, creating a major bottleneck in developing robust models. As an alternative, synthetic data generation using conditional diffusion models such as \cite{rombach2022high} has shown promise in improving downstream performance \citep{yu2023diffusion, oh2023diffmix}.

Conditional diffusion architectures like ControlNet \cite{zhang2023adding} enable image generation from structured inputs such as segmentation masks, edges, or depth maps. In medical imaging, these models generate images conditioned on pixel level annotations \citep{shrivastava2023nasdm}. While such generations often appear plausible and respect spatial layouts, they frequently fail to preserve critical properties such as intensity distributions, texture consistency, and structural correctness. As a result, models trained on such synthetic data may suffer from degraded downstream performance.

\begin{figure}[t]
\centering
\includegraphics[width=\textwidth, height=2.2in]{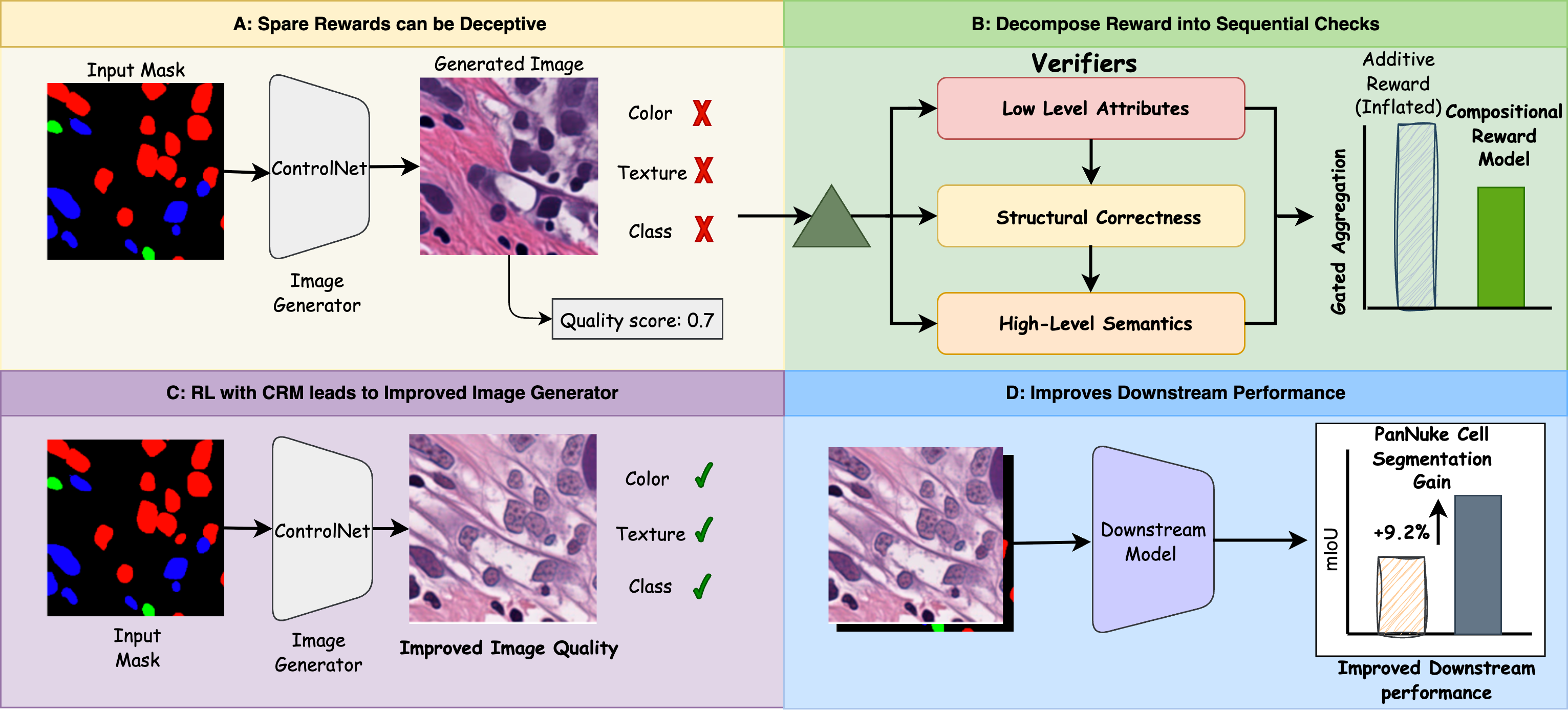}
\caption{A: Conditional generator makes fine grain errors and a single score is not descriptive to provide good feedback. B: We decompose the reward into a composition of rewards (CRM) where, each reward constraint is checked before moving to next reward which we call, hierarchical constraint propagation (HCP). C: Using HCP to train generator with RL improves the image quality and D: eventually improves the performance of downstream task.}
\label{fig:teaser}
\end{figure}

Recent approaches attempt to improve diffusion model outputs by optimizing for perceptual quality using reinforcement learning from AI feedback \citep{black2023training, fan2023reinforcement}. These methods typically rely on an outcome reward model (ORM) that assigns a single scalar score to generated images, often derived from vision language models. However, for medical imaging, such a scalar reward is insufficient, as it conflates multiple aspects of image quality, including intensity fidelity, texture realism, and structural alignment, into a single signal (see Fig \ref{fig:teaser}:A). This lack of granularity limits the effectiveness of learning and provides weak guidance for correcting specific failure modes.

We propose \ours{}, a Compositional Reward Model (CRM) with Hierarchical Constrained Propagation (HCP) framework for optimizing conditional diffusion models. Instead of a single reward, we decompose quality assessment into a set of verifier grounded sub rewards, each corresponding to a distinct stage of evaluation. To ensure that early stages focus on fine grain attributes such as intensity and texture, while later stages assess higher level structural and semantic correctness we use hierarchical constrained propagation (HCP) which enforces ordering: lower level attributes must be satisfied before higher level rewards can be accumulated. This design prevents easier objectives from masking failures in critical components (see Fig \ref{fig:teaser}:B). Finally, we use resulting reward (CRM) to optimize conditional diffusion with reinforcement learning using GDPO \citep{liu2026gdpo}.  


\ours{} not only leads to improved image generation (see Fig.~\ref{fig:teaser}:C) but also improved downstream performance (see Fig.~\ref{fig:teaser}:D). We evaluate \ours{} across three tasks: cell segmentation (PanNuke), villi/crypt detection and measurement (CeDeM), and skin lesion classification (ISIC). Training downstream models on data generated by \ours{} leads to consistent improvements over the closest baseline on each task: a $2.3\%$ increase in mDice for SegFormer \citep{xie2021segformer} on PanNuke, an $8.5\%$ reduction in ratio MRE for MeasureNet \citep{tyagi2024measurenet} on CeDeM, and a $5.9\%$ gain in F1 for ResNet \citep{he2016deep} on ISIC.

In summary, we make following contributions. First, we introduce a Compositional Reward Model (CRM) for conditional diffusion models, which decomposes image quality into verifier grounded sub rewards and combines them through a hierarchical constrained propagation that enforces a hierarchy, enabling fine to coarse correctness. Second, we demonstrate consistent improvements across three medical imaging benchmarks spanning segmentation, measurement, and classification tasks.

\vspace{-1em}

\section{Related Work}
\label{sec:related_work}

\paragraph{Synthetic Medical Image Generation.}

Diffusion models~\citep{ho2020denoising,rombach2022high} have become a standard framework for image synthesis, supporting conditioning through text, class labels, and spatial controls. ControlNet~\citep{zhang2023adding} extends pretrained latent diffusion models with additional structural inputs, making it especially suitable for medical image generation where semantic masks provide pixel level supervision. In medical imaging, mask conditioned diffusion has been explored for nuclei and pathology synthesis~\citep{shrivastava2023nasdm,yu2023diffusion,oh2023diffmix}, anatomy-guided generation~\citep{konz2024anatomically}, and synthetic data augmentation for skin disease classification~\citep{akrout2023diffusion}. These methods improve synthetic data availability, but generated samples may still violate task critical properties such as class specific intensity, texture realism, object count, morphology, or mask-image alignment.

\textbf{Diffusion Models with Reinforcement Learning:}
Recent works have explored reinforcement learning (RL) for aligning diffusion models with downstream objectives. DDPO~\cite{black2023training} formulates the denoising process as a multi-step Markov Decision Process (MDP) and applies policy gradient optimization using reward signals. Subsequent methods such as DPOK \cite{fan2023reinforcement} and DRaFT \cite{prabhudesai2023aligning} introduce KL regularization to stabilize training and mitigate reward hacking. B2-DiffuRL~\cite{hu2025towards} further identifies sparse terminal rewards as a key credit assignment problem in diffusion RL, and mitigates it through backward progressive training and branch based sampling. GRPO~\cite{shao2024deepseekmath} extends this paradigm to multiple rewards but aggregates them in a purely additive manner, which fails to capture dependencies between different quality factors. GDPO~\cite{liu2026gdpo} further improves multi-reward optimization by computing reward level advantages, leading to more stable and effective training. We build on GDPO and demonstrate that incorporating structured, compositional rewards rather than additive combinations leads to significantly improved alignment for conditional medical image generation.

\paragraph{Reward Modeling for Diffusion:}
Once trained, diffusion models can be further aligned with desired objectives using reward based optimization. A common approach relies on inference time selection, where multiple samples are generated for a given input and the best is chosen using a reward function \citep{lightman2023let,stiennon2020learning}. While this improves output quality, it does not correct underlying generation errors. An alternative is to directly optimize the model using reward signals. Existing methods typically employ an output reward model (ORM) \citep{black2023training,cobbe2021training, saremi2025rl4med} that assigns a single scalar score to each generated image, often using vision language models or domain specific evaluators. However, such scalar rewards provide limited supervision, as they collapse multiple aspects of image quality, such as intensity, texture, and structural correctness, into a single signal. Recent work in language models explores Process Reward Models (PRM) \citep{lightman2023let}, which provide intermediate supervision over multi step reasoning. However, PRMs are not directly applicable to diffusion models, where intermediate denoising steps are not semantically meaningful or independently verifiable. In contrast, prior work in VVL/VLM shows that dense and structured reward signals \citep{cui2025process, guo2025can} can significantly improve learning over sparse outcome based rewards. While methods such as GRPO \cite{shao2024deepseekmath} optimize multiple rewards through simple additive aggregation, such formulations fail to capture hierarchical dependencies between different aspects of image quality.

\begin{figure}[t]
\centering
\includegraphics[width=\textwidth, height=2in]{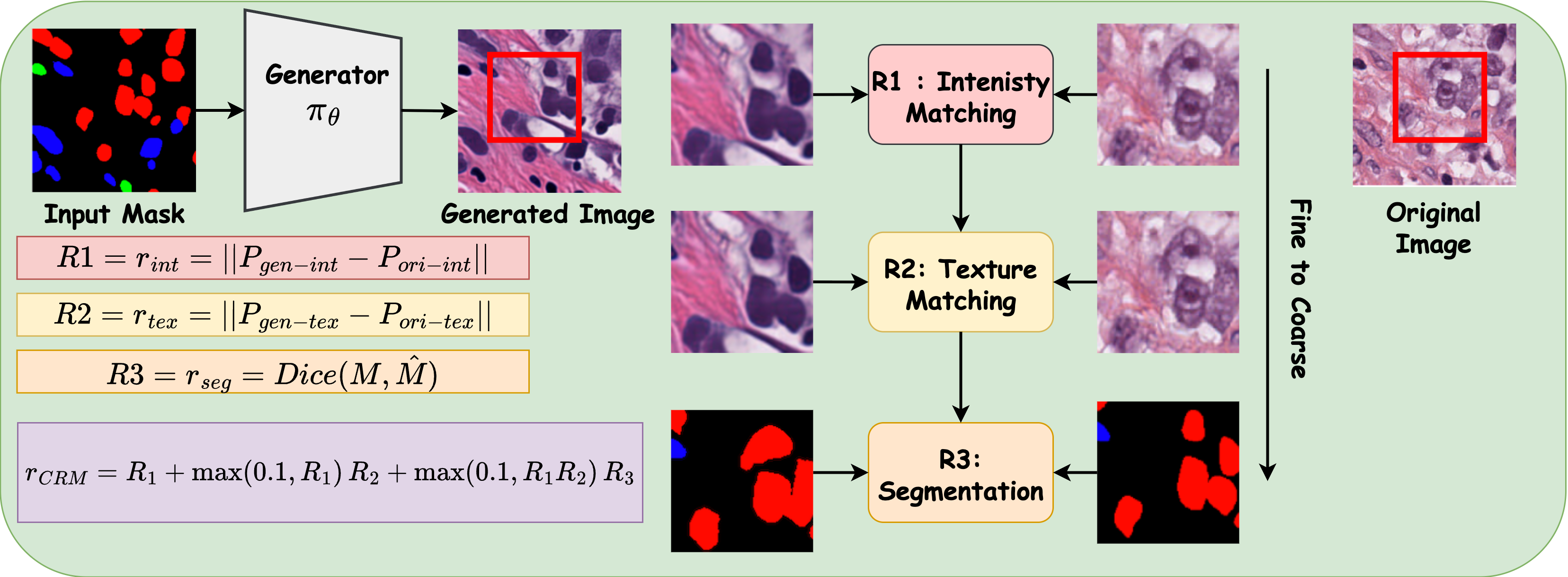}
\caption{Once the generator $\pi_{\theta}$ is trained in a supervised manner, it is used to generate images conditioned on a segmentation mask $M$ and text prompt $T$. Each generated image ($\hat{I}$) is evaluated through set of verifiers. For the PanNuke dataset, these verifiers enforce a fine to coarse hierarchy: moving from low level attributes (intensity and texture matching by cell type) upward to structural and semantic agreement via pre-trained segmentation models. These stage wise sub rewards are composed using Hierarchical Constrained Propagation (HCP), ensuring that lower level constraints must be satisfied before higher level credit can accrue to yield the final gated reward $r_{CRM}$.}
\label{fig:Main_Figure}
\end{figure}
\vspace{-1em}

\section{Method}
\label{sec:method}

Our goal is to generate synthetic data without requiring additional pixel level annotation, to improve downstream models trained on both real and synthetic data. To do so, we use a segmentation mask ($M$) as input to generate an image ($I$) that follows the mask structure. Formally, let $I \in \mathcal{I}$ denote a medical image, $M \in \mathcal{M}$ its semantic mask, and $T \in \mathcal{T}$ a text prompt. In the first stage, we train a conditional diffusion model (ControlNet~\citep{zhang2023adding}) to model the conditional distribution $p_{\theta_0}(I \mid M, T)$ by maximizing the conditional log likelihood:

\begin{equation}
    \max_{\theta_0}\; \mathbb{E}_{(I,M,T)\sim\mathcal{D}}
    \big[\log p_{\theta_0}(I\mid M,T)\big],
    \label{eq:conditional_objective}
\end{equation}

which is implemented via the standard denoising objective~\citep{ho2020denoising}. This stage yields an initial generator $\pi_0=\pi_{\theta_0}$ that produces samples consistent with the conditioning masks. However, this objective is under specified for properties that determine downstream utility, such as fine grain appearance (e.g., intensity and texture), structural alignment with the mask, and high level semantic or morphological correctness (see Fig.~\ref{fig:Main_Figure}). As a result, samples from $\pi_0$ may satisfy coarse conditioning while exhibiting critical fine grain errors along these dimensions.

A common approach to address this limitation is to further align the initial generator using a scalar outcome reward model (ORM), which assigns a single reward to each generated sample. We instead propose \ours{}, a Compositional Reward Model (CRM) framework that decomposes image quality into verifier grounded reward components, each targeting a specific attribute. For example, in PanNuke we evaluate cell intensity consistency and nuclei texture along with semantic structure (see Fig.~\ref{fig:Main_Figure}), in CeDeM we enforce villi shoulder and crypt border alignment along with morphological consistency (see Fig~\ref{app:dataset_examples}), and in ISIC we assess lesion intensity, mask agreement, and diagnostic class correctness (see Fig~\ref{app:dataset_examples}) . Each component is evaluated by a dedicated verifier and captures a distinct aspect of correctness, including appearance fidelity, structural consistency, morphological validity, and semantic alignment. During RL alignment, $\pi_\theta$ denotes the current policy initialized from $\pi_0$. Formally, for a generated sample $\hat{I} \sim \pi_\theta(\cdot \mid M, T)$, we define a set of rewards $\{r_k(\hat{I}, M, T)\}_{k=1}^K$, where each $r_k$ corresponds to a specific quality criterion.


Instead of aggregating these rewards additively, \ours{} composes them through Hierarchical Constrained Propagation (HCP), enforcing an ordering over the verification objectives. In particular, low level attributes (e.g., intensity and texture) act as prerequisites for high level rewards (e.g., structural or semantic correctness), ensuring that failures in fundamental properties are not masked by success in easier objectives. The resulting reward is used to optimize $\pi_\theta$ via reinforcement learning. We adopt Group Distributional Policy Optimization (GDPO)~\citep{liu2026gdpo}, which enables stable training under multiple reward signals through group comparisons and decoupled component normalization.

\subsection{Compositional Reward Model}
\label{subsec:crm}

The Compositional Reward Model (CRM) combines $K$ bounded sub rewards
$\{r_k(\mathbf{x}_0,\mathbf{c}) \in [0,1]\}$ into a single scalar
$r_\mathrm{CRM} \in [0,1]$. Each component targets a distinct,
verifiable failure mode of the generator, computed using frozen
verifiers. For example, in cell histopathology (PanNuke), separate
rewards assess whether the generated cells exhibit correct
class specific intensity distributions, realistic texture patterns,
and semantic consistency with the conditioning mask. Similar
task specific decompositions are used for CeDeM (alignment,
morphology, and ratio) and ISIC (lesion appearance, attributes,
and diagnostic class), as illustrated in Fig.~\ref{fig:Main_Figure}.
By explicitly separating these factors, CRM provides structured
feedback that is aligned with the requirements of downstream tasks.


Instead of aggregating rewards additively, CRM uses hierarchical
constrained propagation, where later components are gated by the success
of earlier ones. This enforces a hierarchical structure: low level
attributes must be satisfied before higher level rewards can contribute.
Let $\gamma=0.1$ denote a minimum gate to preserve gradient signal. The
composed reward is defined as

\begin{equation}
\begin{aligned}
r_\mathrm{CRM}(\mathbf{x}_0,\mathbf{c})
&= \frac{1}{K}\sum_{k=1}^{K} r_k(\mathbf{x}_0,\mathbf{c})\,G_k(\mathbf{x}_0,\mathbf{c}), \\
G_1(\mathbf{x}_0,\mathbf{c}) &= 1, \\
G_k(\mathbf{x}_0,\mathbf{c})
&= \max\!\left(\gamma,\prod_{\ell<k} r_\ell(\mathbf{x}_0,\mathbf{c})\right),\quad k>1,
\end{aligned}
\label{eq:cascading_reward}
\end{equation}

Here, each gate $G_k$ modulates the contribution of $r_k$ based on the
cumulative success in hierarchical structure. The minimum gate $\gamma$
prevents a single failed component from collapsing the reward to zero.
The composition is order sensitive; and follows a fine
to coarse progression (see Sec.~\ref{sec:results}).

\subsection{Rewards}
\label{subsec:sub_rewards}

CRM is built on two design principles: (i) each reward component is
verifiable, bounded in $[0,1]$, and computed using a frozen model or
a closed form statistic; (ii) rewards follow a prerequisite ordering,
where lower level attributes (e.g., intensity and texture) are
evaluated before higher level semantic properties.
Across datasets, we instantiate three classes of rewards:
(i)~distributional matching, (ii)~segmentation agreement, and
(iii)~classification based scores.

\textbf{Distributional matching.}
We model appearance and texture using class conditional Gaussian
mixture models (GMMs)
$P=\sum_{k=1}^{K}\phi_k\mathcal{N}(u;\mu_k,\Sigma_k)$,
and compare features $u_i$ extracted from generated samples against
a reference likelihood
$\ell^\star=\mathbb{E}_{u\sim P}\log P(u)$:
\begin{equation}
R_\mathrm{GMM}(\mathcal{X};P)
= \exp\!\Big(-\big(\ell^\star - \tfrac{1}{n}\!\sum_{i}\log P(u_i)\big)_{+}\big/\tau\Big)\in(0,1],
\label{eq:gmm_reward}
\end{equation}
which follows the DeGPR style class conditional regularization~\cite{tyagi2023degpr}.

\textbf{Mask agreement.}
We measure structural consistency using the Dice score:
\begin{equation}
\mathrm{Dice}(M,\hat{M})=\frac{2|M\cap \hat{M}|}{|M|+|\hat{M}|}\in[0,1],
\label{eq:dice_reward}
\end{equation}

\textbf{Classification score.}
We use the softmax margin of a frozen classifier $\psi$ with logits
$\boldsymbol{\ell}=\psi(\mathbf{x}_0)$ for target class $y$:
\begin{equation}
M_y(\mathbf{x}_0) = \sigma\!\Big(\ell_y - \max_{c\neq y}\ell_c\Big)\in(0,1).
\label{eq:margin}
\end{equation}

\paragraph{PanNuke.}
PanNuke~\citep{gamper2020pannuke} contains six class nuclei
segmentation pairs, with downstream task of multi class segmentation.
CRM enforces that generated nuclei match class specific appearance
before receiving segmentation level credit.
For foreground classes $c\in\mathcal{C}_f$, pixel sets $\mathcal{X}_c$,
and texture descriptors $\mathcal{T}_c$, the reward order is
intensity $\rightarrow$ texture $\rightarrow$ segmentation:
\begin{equation}
\begin{aligned}
r_\mathrm{int}
&= \frac{1}{|\mathcal{C}_f|}\sum_{c\in\mathcal{C}_f}
R_\mathrm{GMM}(\mathcal{X}_c;P_c), \\
r_\mathrm{tex}
&= \frac{1}{|\mathcal{C}_f|}\sum_{c\in\mathcal{C}_f}
R_\mathrm{GMM}(\mathcal{T}_c;Q_c), \\
r_\mathrm{seg}
&= \mathrm{Dice}(\phi(\mathbf{x}_0),\mathbf{m}).
\end{aligned}
\label{eq:pannuke_rewards}
\end{equation}
Here $\phi$ denotes a frozen SegFormer-B3~\citep{xie2021segformer}.

\paragraph{CeDeM.}
CeDeM~\citep{tyagi2024measurenet} provides duodenal biopsy images in
which villi and crypts are annotated by bisector polylines, with
additional villus shoulder (VS) and crypt border (CB) annotations.
The downstream model, MeasureNet, frames measurement as polyline
detection: the detection branch predicts villus and crypt polylines. Additional segmentor $\phi$ is use to predict segmentation masks for villi shoulder and crypt border. We use the detection output for evaluation. Accurate localization of villi shoulder anchors is the dominant
failure mode.

\emph{Anchor consistency.}
We enforce alignment using a frozen segmenter $\phi$ and compute
segmentation agreement $d_{\mathrm{seg}}$ and contour matching
via Dynamic Time Warping (DTW) distance $d_{\mathrm{dtw}}$:
\begin{equation}
r_{\mathrm{anchor}} =
\lambda_{\mathrm{seg}}
\exp\!\left(-\frac{d_{\mathrm{seg}}^2}{2\sigma_{\mathrm{seg}}^2}\right)
+
\lambda_{\mathrm{dtw}}
\exp\!\left(-\frac{d_{\mathrm{dtw}}^2}{2\sigma_{\mathrm{dtw}}^2}\right),
\label{eq:cedem_Anchor}
\end{equation}

\emph{Morphological consistency.}
We measure geometric alignment between predicted and reference
polylines by MeasureNet \citep{tyagi2024measurenet} using Earth Mover’s Distance (EMD):
\begin{equation}
r_{\mathrm{morph}} =
\exp\!\left(-\frac{d_{\mathrm{emd}}^2}{2\sigma_{\mathrm{emd}}^2}\right).
\label{eq:cedem_morphlogy}
\end{equation}

\emph{Ratio consistency.}
Finally, we enforce agreement with the target villi-to-crypt ratio
$\rho^\star$:
\begin{equation}
r_{\mathrm{ratio}} =
\exp\!\left(-\frac{(\hat{\rho} - \rho^\star)^2}{2\sigma_{\rho}^2}\right).
\label{eq:cedem_ratio}
\end{equation}

The resulting ordering is
anchor $\rightarrow$ morphology $\rightarrow$ ratio.

\paragraph{ISIC.}
ISIC~\citep{codella2019skin} conditions on a lesion mask and a
diagnostic class token $y$ (e.g., \texttt{melanoma}), with the
downstream task being lesion classification. A key challenge is that
optimizing only for classifier confidence can produce unrealistic
images that exploit spurious cues. To mitigate this, we decompose the
reward into appearance, attribute, and diagnostic components.

Let $\mathcal{X}_{\mathrm{les}}$ denote lesion pixels,
$P_{\mathrm{les}}^{(y)}$ a class conditional GMM capturing lesion
appearance, $\phi_{\mathrm{les}}$ a frozen lesion segmentation
model and $M_{y}$ being diagnosis classifier. We define:
\begin{equation}
\begin{aligned}
r_{\mathrm{int}} 
&= R_{\mathrm{GMM}}(\mathcal{X}_{\mathrm{les}}; P_{\mathrm{les}}^{(y)}), \\
r_{\mathrm{attr}} 
&= \mathrm{Dice}(\phi_{\mathrm{les}}(\mathbf{x}_0), \mathbf{m}_{\mathrm{les}}), \\
r_{\mathrm{dx}} 
&= M_y(\mathbf{x}_0).
\end{aligned}
\label{eq:isic_rewards}
\end{equation}

The rewards are applied in the order
intensity $\rightarrow$ attribute mask $\rightarrow$ diagnostic class,
ensuring that diagnostic credit is assigned only when appearance and
lesion structure are consistent.

\subsection{Optimization Generator with GDPO}
\label{subsec:controlnet_gdpo}

For RL finetuning, we denote the current diffusion policy as $\pi_\theta$,
initialized from the pretrained model $\pi_0$. Given a conditioning input
$\mathbf{c}=\mathbf{c}(M,T)$, the policy generates a sample via the
denoising trajectory, producing a final image from latent
$\mathbf{x}_0$. Following~\cite{black2023training}, denoising is modeled as a multi step
MDP with state $\mathbf{s}_t=(\mathbf{x}_t,\mathbf{c},t)$ and action
$\mathbf{a}_t=\boldsymbol{\epsilon}_\theta(\mathbf{x}_t,\mathbf{c},t)$,
with a terminal reward at $\mathbf{x}_0$. We use the CRM reward vector
$\mathbf{r}(\mathbf{x}_0,\mathbf{c})=(r_1,\dots,r_K)$ and optimize the
policy using GDPO~\cite{liu2026gdpo}, which normalizes each component
before aggregation. For each conditioning $\mathbf{c}_i$, we sample $N$ candidates
$\{\mathbf{x}_{0,i}^{(j)}\}_{j=1}^{N}$ and form the HCP gated components
$\tilde{r}_{i,k}^{(j)}=r_{i,k}^{(j)}\,G_k^{(j)}$, on which GDPO operates:

\begin{equation}
\tilde{A}_{i,k}^{(j)} = \frac{\tilde{r}_{i,k}^{(j)} - \mu_{i,k}}{\sigma_{i,k}+\epsilon}, \quad
\hat{A}_i^{(j)} = \sum_{k} w_k\,\tilde{A}_{i,k}^{(j)}, \quad
A_i^{(j)} = \frac{\hat{A}_i^{(j)} - \bar{\hat A}_i}{\sigma_{\hat A_i}+\epsilon},
\label{eq:gdpo_advantage}
\end{equation}

Let $\pi_{\theta_\mathrm{old}}$ denote the rollout policy. GDPO updates
$\pi_\theta$ using the likelihood ratio and rollout policy defines a PPO style trust region~\cite{schulman2017proximal}.
\begin{equation}
\rho_{i,j,t}(\theta) =
\exp\!\left(
\log \pi_\theta(\mathbf{a}_{i,j,t}\mid\mathbf{s}_{i,j,t}) -
\log \pi_{\theta_\mathrm{old}}(\mathbf{a}_{i,j,t}\mid\mathbf{s}_{i,j,t})
\right),
\label{eq:gdpo_ratio}
\end{equation}
and the clipped surrogate objective
\begin{equation}
\mathcal{L}_\mathrm{GDPO}(\theta)
= -\mathbb{E}_{i,j,t}\Big[
\min\big(
\rho_{i,j,t}(\theta) A_i^{(j)},
\mathrm{clip}(\rho_{i,j,t}(\theta),1-\epsilon_\mathrm{clip},1+\epsilon_\mathrm{clip})
A_i^{(j)}\big)
\Big],
\label{eq:gdpo_loss}
\end{equation}

\section{Experiments}
\label{sec:experiments}

\paragraph{Datasets and Tasks}
\label{subsec:datasets}
We evaluate \ours{} on three medical imaging benchmarks spanning
segmentation, measurement, and classification.

\paragraph{PanNuke.}
PanNuke~\citep{gamper2020pannuke} is a nuclei segmentation dataset H\&E patches and pixel level labels for five classes (neoplastic,
inflammatory, connective, dead, non neoplastic epithelial).  We use $1143/245/246$ samples for train/val/test. The downstream task is semantic segmentation using SegFormer-B3~\citep{xie2021segformer}(for details \ref{sec:supp_downstream_segformer_pannuke}). 
\textbf{CeDeM} ~\citep{tyagi2024measurenet} is a duodenum biopsy
histopathology dataset for celiac disease measurement. Villi and crypts are annotated by bisector polylines, with additional villus shoulder (VS) and crypt border (CB) annotations used to construct the conditioning maps.
The downstream task is villus-to-crypt length ratio computation with MeasureNet~\citep{tyagi2024measurenet}. We use $1169$ samples for training, $146$ for validation, and $146$ for test (more details \ref{sec:supp_downstream_measurenet}). \textbf{ISIC} ~\citep{codella2019skin} is a public dermoscopy benchmark for skin lesion analysis. We use the ISIC-2019 ~\citep{codella2019skin} for multiclass diagnosis (e.g.\ melanoma, nevus, basal cellcarcinoma). We use $2339$ images for training, $252$ for validation, and $1297$ for test. The downstream task is multiclass image classification with a ResNet-50 backbone~\citep{he2016deep}. Representative conditioning masks and images for all three datasets are shown in ~\ref{app:dataset_examples}.

\paragraph{Implementation Details}
\label{subsec:implementation}

We ControlNet~\citep{zhang2023adding} initialized from Stable Diffusion~2.1~\citep{rombach2022high} and
trained at $512{\times}512$ resolution for conditional generation. We first run supervised ControlNet training with the standard diffusion noise prediction
loss. We optimise with AdamW \citep{loshchilov2017decoupled}  at learning rate (lr)
$10^{-5}$, batch size $4$ per GPU, and image resolution $512{\times}512$, for up to $250$ epochs. The best checkpoint $\pi_0$ is selected by lowest
validation FID, and GDPO finetuning is launched from the same $\pi_0$ for every reward variant. To construct synthetic datasets, we sample from $\pi_0$ using
augmented conditioning masks. Augmentations include geometric
transforms (rotation, flip, translation), structural perturbations
(e.g., random removal of cells or villi), and scale variations
(zoom in/out for ISIC). Further details are provided in
Appendix~\ref{app:augmentation}. For RL finetuning using GDPO, we use AdamW optimizer, lr of $10^{-6}$, group size 32, clip ratio 0.1, inner iterations 2 and DDIM sampling steps 100 (more details Tab. \ref{tab:gdpo_hparams}). For downstream model we use ISIC - ResNet50 \citep{he2016deep}, lr $10^{-4}$, batch size 64 , PanNuke - Segformer \citep{xie2021segformer} with lr 6X$10^{-5}$ and MeasureNet \citep{tyagi2024measurenet} with lr $10^{-4}$, more details (see \ref{sec:downstream}). All experiments were conducted on 80GB A100 GPUs. RL fine tuning of generator (ControlNet) with GDPO uses 8×A100 GPUs and requires approximately 48 hours, followed by 4–8 hours for sample generation, depending on the dataset and downstream training times of 6 hours for ResNet, 10 hours for SegFormer, and 18 hours for MeasureNet.

\paragraph{Evaluation Metrics}
\label{subsec:metrics}
For PanNuke, we report mean IoU and mean Dice. For CeDeM, we report mean absolute error (MAE) and mean relative error (MRE) for the villus, crypt, villi-to-crypt ratio. For ISIC, we report accuracy and macro F1, with more details provided in Appendix~\ref{app:full_metrics}.

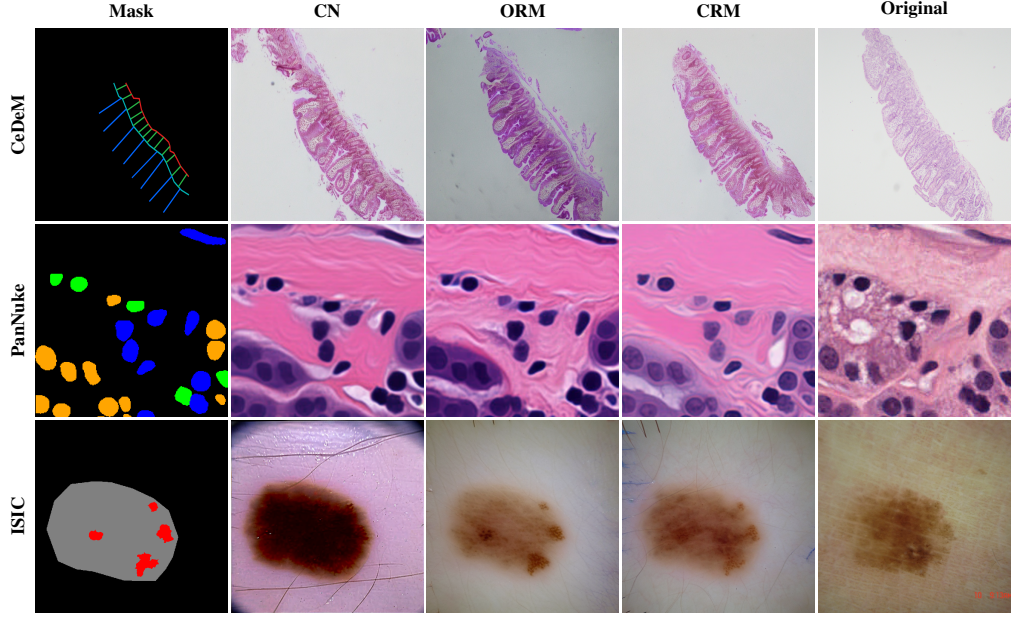
\begin{figure}[t]
\centering
\begin{tikzpicture}[
    picture format/.style={inner sep=0.5pt},
]
\node[picture format] (A1)
    {\includegraphics[width=1in,height=1in]{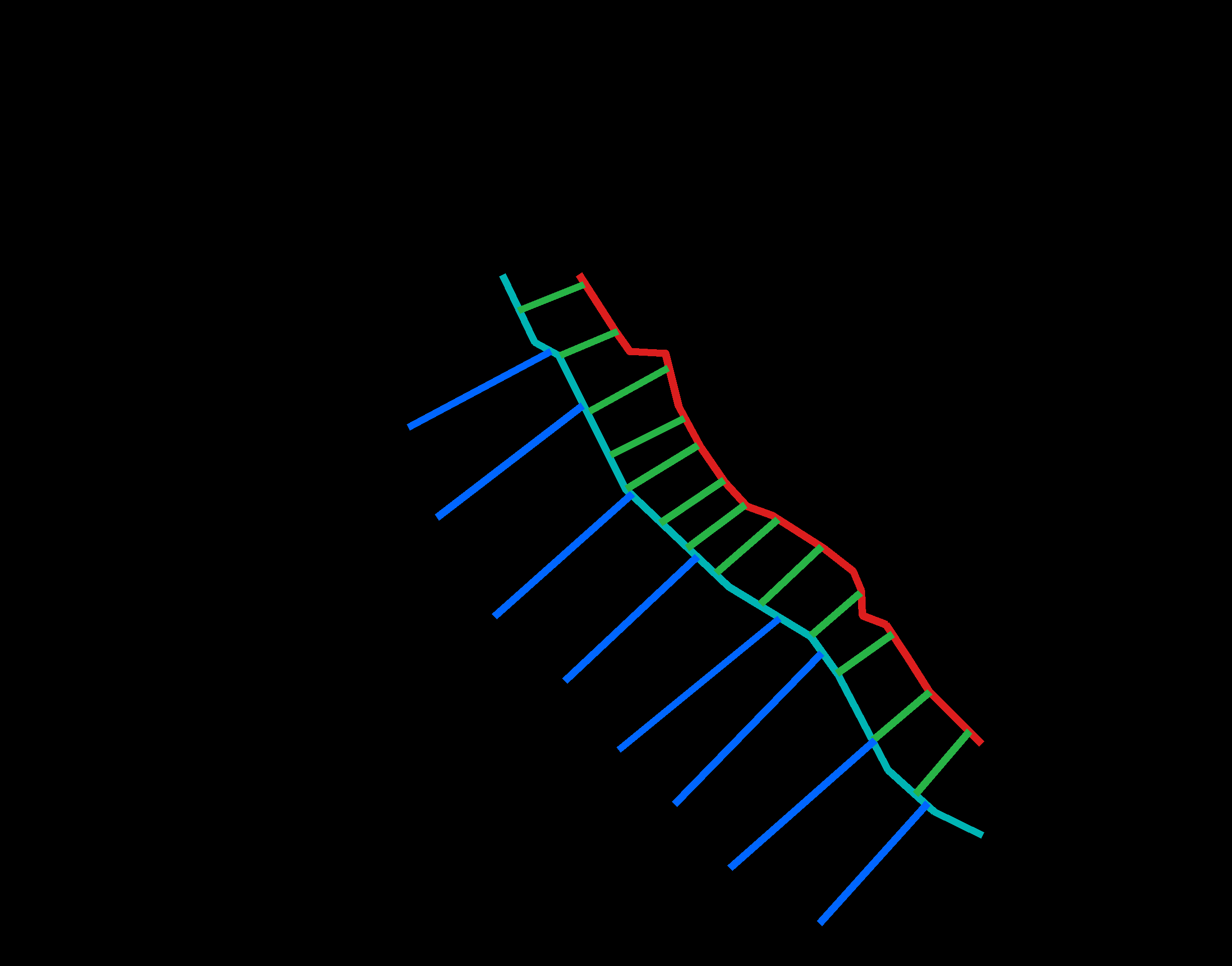}};
\node[picture format, anchor=north west] (A2) at (A1.north east)
    {\includegraphics[width=1in,height=1in]{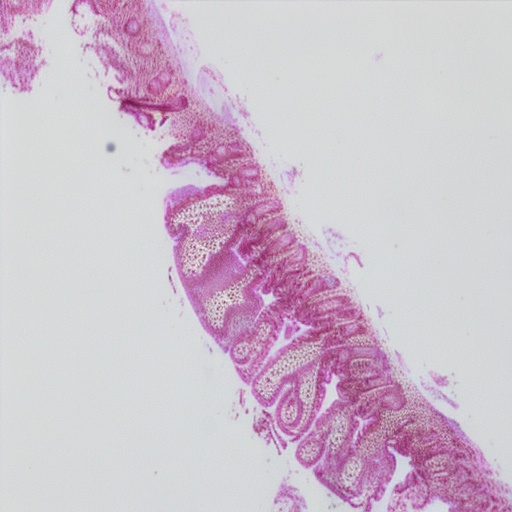}};
\node[picture format, anchor=north west] (A3) at (A2.north east)
    {\includegraphics[width=1in,height=1in]{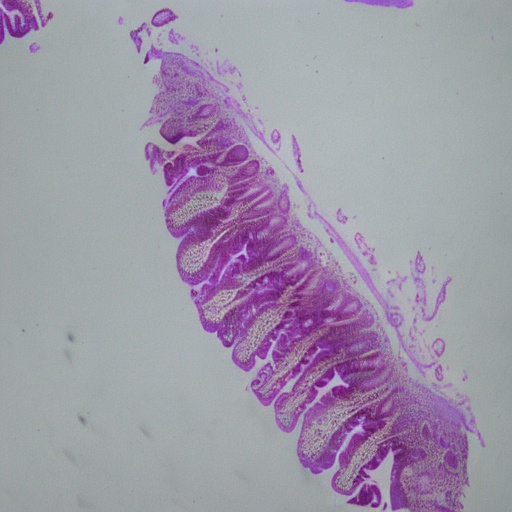}};
\node[picture format, anchor=north west] (A4) at (A3.north east)
    {\includegraphics[width=1in,height=1in]{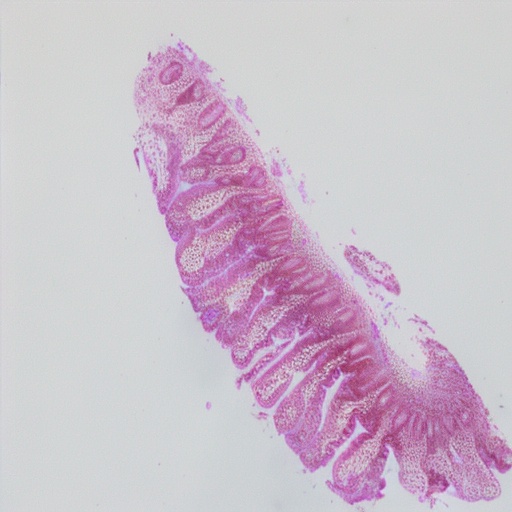}};
\node[picture format, anchor=north west] (A5) at (A4.north east)
    {\includegraphics[width=1in,height=1in]{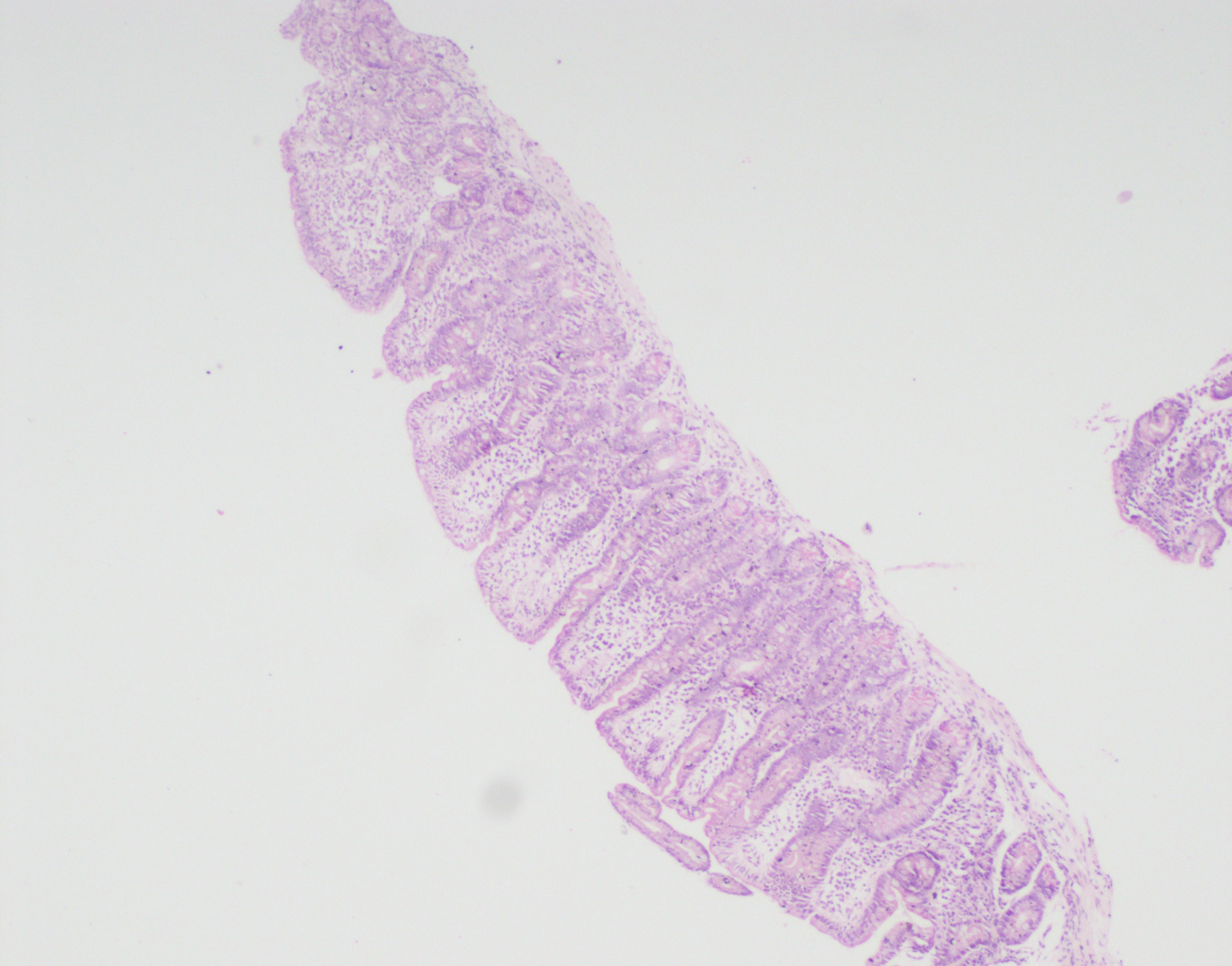}};

\node[picture format, anchor=north] (B1) at (A1.south)
    {\includegraphics[width=1in,height=1in]{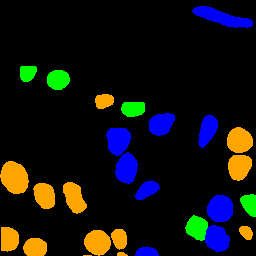}};
\node[picture format, anchor=north] (B2) at (A2.south)
    {\includegraphics[width=1in,height=1in]{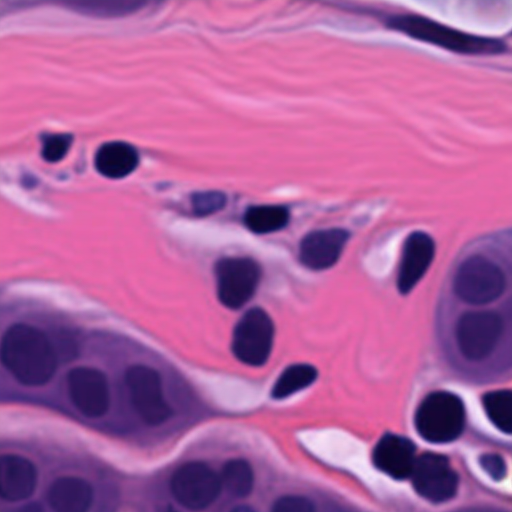}};
\node[picture format, anchor=north] (B3) at (A3.south)
    {\includegraphics[width=1in,height=1in]{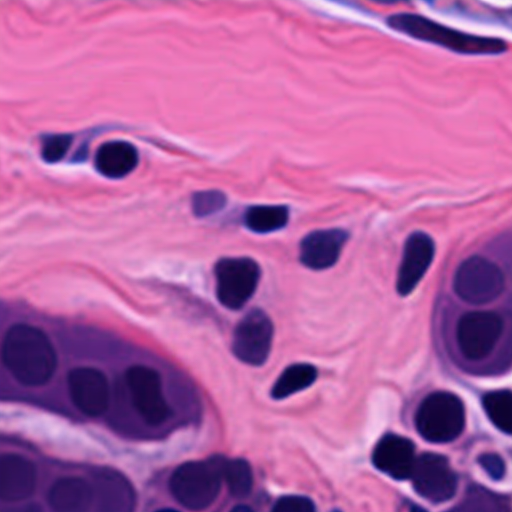}};
\node[picture format, anchor=north] (B4) at (A4.south)
    {\includegraphics[width=1in,height=1in]{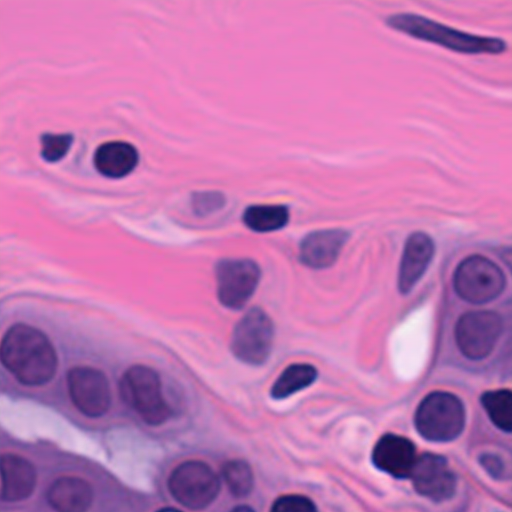}};
\node[picture format, anchor=north] (B5) at (A5.south)
    {\includegraphics[width=1in,height=1in]{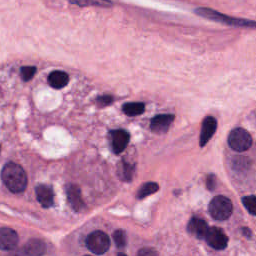}};

\node[picture format, anchor=north] (C1) at (B1.south)
    {\includegraphics[width=1in,height=1in]{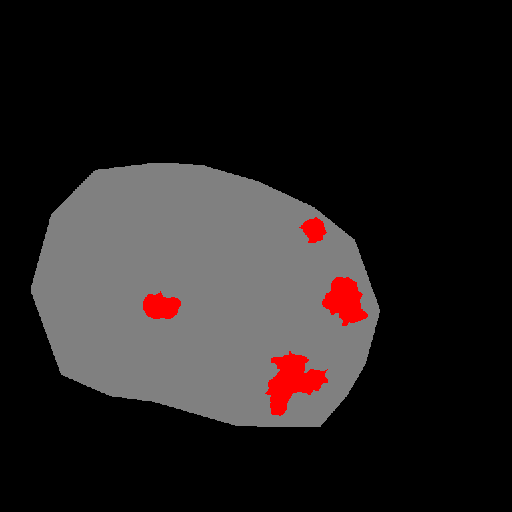}};
\node[picture format, anchor=north] (C2) at (B2.south)
    {\includegraphics[width=1in,height=1in]{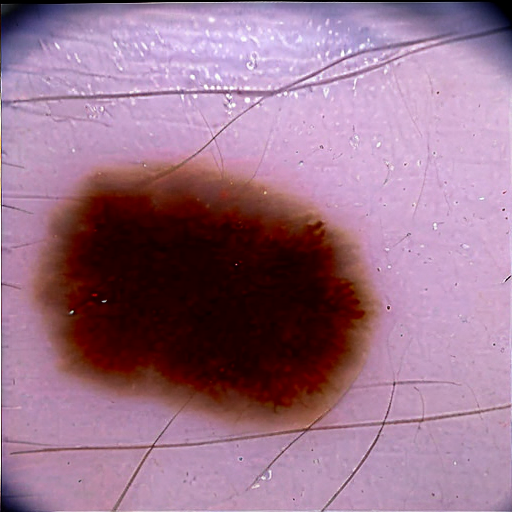}};
\node[picture format, anchor=north] (C3) at (B3.south)
    {\includegraphics[width=1in,height=1in]{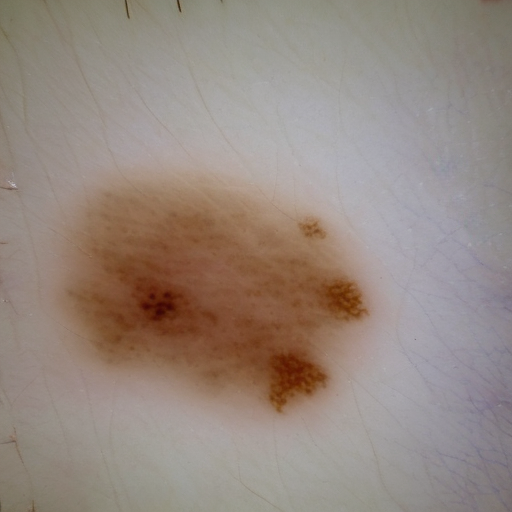}};
\node[picture format, anchor=north] (C4) at (B4.south)
    {\includegraphics[width=1in,height=1in]{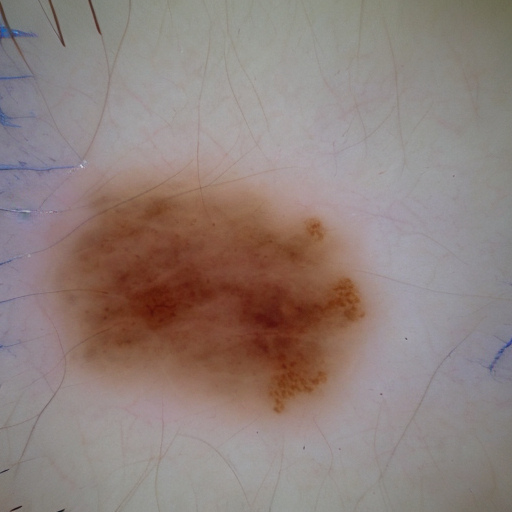}};
\node[picture format, anchor=north] (C5) at (B5.south)
    {\includegraphics[width=1in,height=1in]{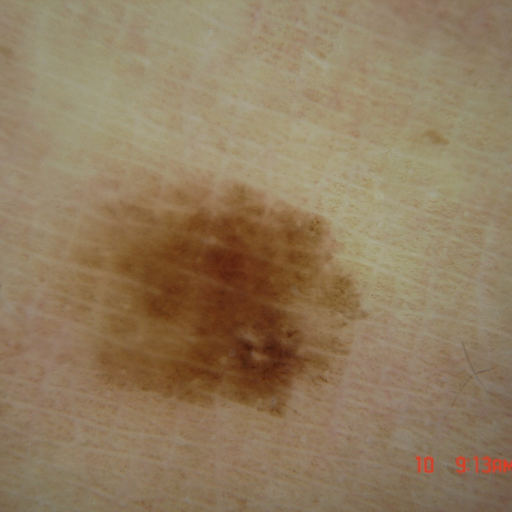}};

\node[anchor=south] at (A1.north) {\scriptsize \textbf{Mask}};
\node[anchor=south] at (A2.north) {\scriptsize \textbf{CN}};
\node[anchor=south] at (A3.north) {\scriptsize \textbf{ORM}};
\node[anchor=south] at (A4.north) {\scriptsize \textbf{CRM}};
\node[anchor=south] at (A5.north) {\scriptsize \textbf{Original}};

\node[rotate=90, anchor=south] at (A1.west) {\scriptsize \textbf{CeDeM}};
\node[rotate=90, anchor=south] at (B1.west) {\scriptsize \textbf{PanNuke}};
\node[rotate=90, anchor=south] at (C1.west) {\scriptsize \textbf{ISIC}};

\end{tikzpicture}
    \caption{Qualitative comparisons for CeDeM, PanNuke, and ISIC.
    Each row uses the same conditioning mask across the supervised
    ControlNet (CN), ORM, and CRM generations. The original image is
    shown only as the paired real target for reference.}
    \label{fig:qualitative_figure}
\end{figure}

\section{Results}
\label{sec:results}

\paragraph{Main quantitative and qualitative comparison.}
Table~\ref{tab:main_results} summarizes downstream performance across all three tasks. 
Augmenting the original training data with synthetic samples from $\pi_0$ consistently improves performance (e.g., $-4.37$ MAE$_\text{R}$ on CeDeM, $+1.90$ mIoU on PanNuke, and $+2.39$ accuracy on ISIC), confirming that ControlNet is an effective data augmenter. Best-of-N: Sampling N samples and picking best sample based on ORM further improves the downstream performance. RL based alignment provides further gains only 
when the reward is well structured: while scalar rewards (ORM) yield moderate improvements and additive compositions SUM or MaxMin (where we maximize the minimum reward) are inconsistent, CRM achieves the best performance 
across all datasets, with $39.58$ MAE$_\text{R}$ on CeDeM, $61.54$ mIoU on PanNuke, and $64.00$ accuracy on ISIC. 
Compared to training on original data only, this corresponds to a $24.0\%$ reduction in MAE$_\text{R}$ on CeDeM, 
$+5.18$ mIoU on PanNuke, and $+10.57$ accuracy points on ISIC, while also outperforming the strongest RL baselines in each case.
Additional metrics are provided in Appendix~\ref{app:full_metrics}
These improvements stem from CRM’s structured reward design.  
This is particularly beneficial for tasks sensitive to fine grained errors, such as CeDeM (where Villi Shoulder impacts the length) and ISIC. 
Qualitatively (Fig.~\ref{fig:qualitative_figure}), ORM based samples often preserve coarse appearance but distort fine structures (e.g., villus shoulder and crypt boundaries in CeDeM, nuclei texture in PanNuke, and lesion borders in ISIC), whereas CRM samples better align with conditioning masks while maintaining 
realistic appearance (See Appendix~\ref{sec:supp_qualitative}). We repeat each downstream experiment three times per method and perform paired two sided $t$ tests against the strongest baseline, observing statistically significant improvements for CRM on CeDeM, PanNuke, and ISIC ($p < 0.001$ for all).

\begin{table}[t]
\centering
\caption{Main results across all three datasets and downstream
tasks. CeDeM uses MeasureNet~\citep{tyagi2024measurenet} (lower
is better); PanNuke uses SegFormer-B3~\citep{xie2021segformer}
(higher is better); ISIC uses ResNet-50~\citep{he2016deep}
(higher is better). Best in \textbf{bold}, second best
\underline{underlined}. Full per dataset metrics are in
Appendix~\ref{app:full_metrics}.}
\label{tab:main_results}
\resizebox{\textwidth}{!}{%
\begin{tabular}{@{}ll cc cc cc@{}}
\toprule
& & \multicolumn{2}{c}{\textbf{CeDeM} (MeasureNet)} & \multicolumn{2}{c}{\textbf{PanNuke} (SegFormer-B3)} & \multicolumn{2}{c}{\textbf{ISIC} (ResNet-50)} \\
\cmidrule(lr){3-4} \cmidrule(lr){5-6} \cmidrule(lr){7-8}
\textbf{Method} & \textbf{Reward} & MAE$_\text{R}\downarrow$ & MRE$_\text{R}\downarrow$ & mIoU$\uparrow$ & mDice$\uparrow$ & Acc.$\uparrow$ & F1$\uparrow$ \\
\midrule
SFT (Original only)              & --                                & 52.06          & 24.77          & 56.36          & 40.94          & 53.43          & 44.84 \\
Generated (SFT)                  & --                                & 47.69          & 22.15          & 58.26          & 42.24          & 55.82          & \underline{47.30} \\
Best-of-N                        & Inference                         & 45.71          & 21.83          & 58.61            & 43.09            & 56.00            & 46.85 \\
\midrule
GDPO~\cite{liu2026gdpo}          & ORM                               & \underline{41.95} & \underline{18.85} & 58.88          & 43.27          & \underline{63.00} & 45.50 \\
GDPO~\cite{liu2026gdpo}          & SUM                               & 45.33          & 21.41          & \underline{60.50} & \underline{44.75} & 59.80          & 43.80 \\
GDPO~\cite{liu2026gdpo}          & MaxMin                            & 47.07          & 21.55          & 59.08          & 43.70          & 57.70          & 44.90 \\
\rowcolor{blue!15}
\textbf{GDPO (ours)} & \textbf{CRM} & \textbf{39.58} & \textbf{17.24} & \textbf{61.54} & \textbf{45.77} & \textbf{64.00} & \textbf{50.10} \\
\bottomrule
\end{tabular}
}
\vspace{-1em}
\end{table}

\begin{wraptable}{r}{0.5\textwidth}
\vspace{-2em}
\centering
\scriptsize
\setlength{\tabcolsep}{3pt}
\renewcommand{\arraystretch}{1.05}
\caption{Subreward ablation on CeDeM (MeasureNet, GDPO).
$R_1$: anchor, $R_2$: detection, $R_3$: ratio. Lower is better.}
\label{tab:cedem_subreward}
\begin{tabular}{ccc cc cc cc}
\toprule
$R_1$ & $R_2$ & $R_3$ &
\multicolumn{2}{c}{Villi} &
\multicolumn{2}{c}{Crypt} &
\multicolumn{2}{c}{Ratio} \\
\cmidrule(lr){4-5}\cmidrule(lr){6-7}\cmidrule(lr){8-9}
 & & & MAE & MRE & MAE & MRE & MAE & MRE \\
\midrule
            &            & \checkmark & 11.39 & 10.98 & 7.02 & 15.67 & 41.95 & 18.85 \\
            & \checkmark &            & 11.51 & 10.70 & 7.64 & 18.03 & 48.46 & 22.22 \\
\checkmark  &            &            & 10.02 & 10.54 & 7.76 & 19.84 & 46.00 & 20.72 \\
            & \checkmark & \checkmark &  \textbf{9.31} &  9.70 & 7.39 & 17.67 & 41.87 & 18.78 \\
\checkmark  & \checkmark & \checkmark & 10.01 & \textbf{9.68} & \textbf{6.64} & \textbf{14.43} & \textbf{39.58} & \textbf{17.24} \\
\bottomrule
\end{tabular}
\vspace{-1em}
\end{wraptable}

\subsection{Ablation Studies}
\label{subsec:ablations}

\textbf{Subreward components:}
Table~\ref{tab:cedem_subreward} ablates the three CRM components on
CeDeM. With only $R_3$ (ratio), model not only misses to predict correct villi shoulder (starting point of villi),
but also make additional partial villis which leads to false positive villi predictions. With only $R_2$ (detection), 
the model learns to predict the correct number of villi and crypts but still makes error regarding the starting point of villi. 
With only $R_1$ (anchor), the model learns to place villi and crypts in the right locations, but makes error in detection. 
With CRM based reward composition, for $R_2$ and $R_3$, improves MAE$\text{R}$ to 41.87. Futher adding $R_1$, shows a improvement of $-0.17$ MAE$_\text{R}$, confirming each reward contribute is essential.

\begin{figure}[t]
    \centering
    \includegraphics[width=\textwidth]{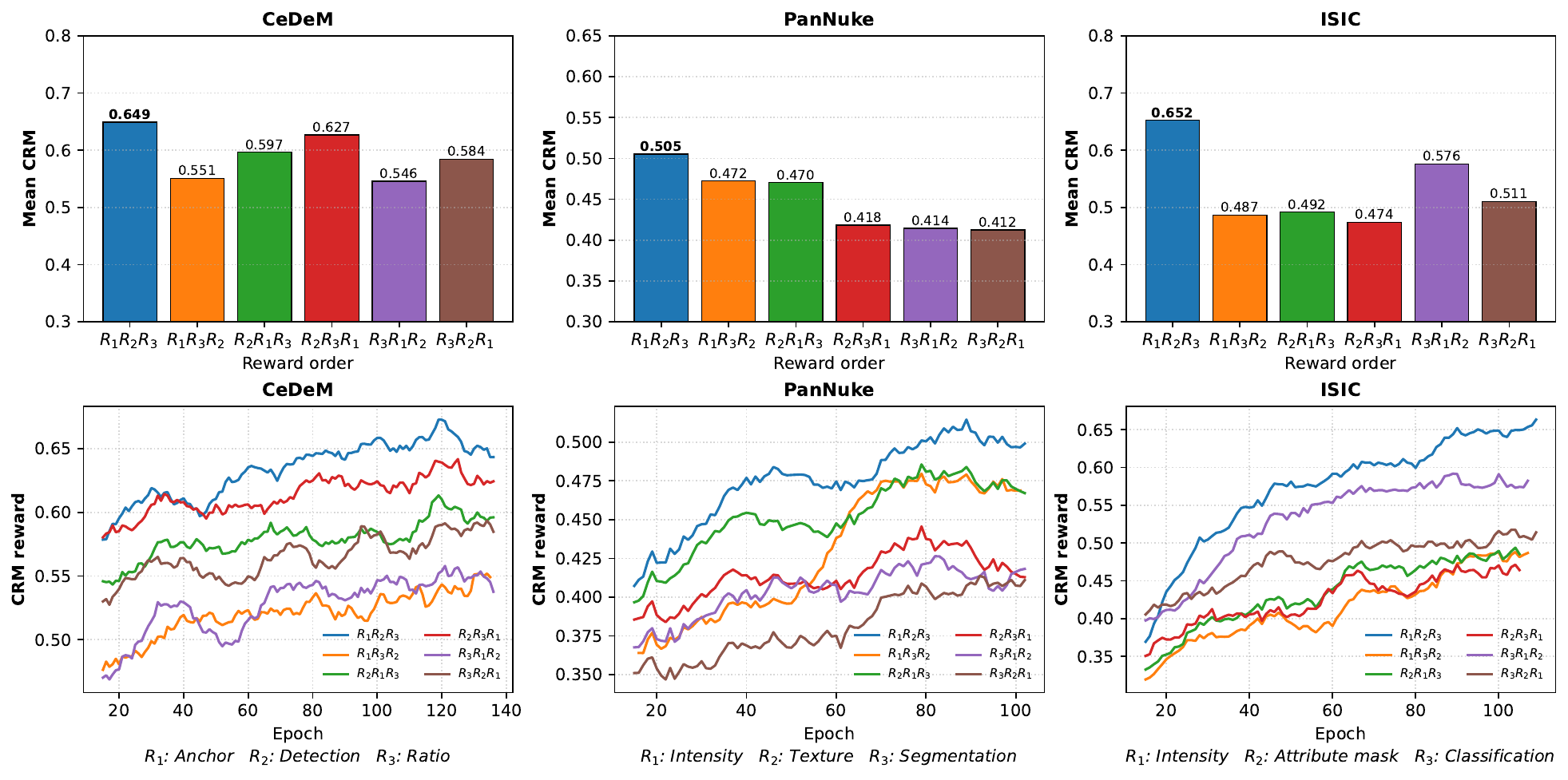}
    \caption{CRM order ablation on CeDeM, PanNuke, and
    ISIC using the available permutation logs. The top row shows
    the mean CRM for last 15 epochs; the bottom row shows
    moving average training trajectories.}
    \label{fig:permute_crm_order}
\end{figure}

\paragraph{Sensitivity to CRM reward order.}
CRM is not order invariant: changing the order changes
which later rewards are gated by earlier successes. We therefore
permute the reward groups while keeping the components and training
budget fixed, and compare the resulting training rewards in
Fig.~\ref{fig:permute_crm_order}. In all the cases, the canonical order of $R_1{\to}R_2{\to}R_3$ is able to extract maximum reward
, where R3 is used first, the reward is not able to extract the signal from R1 and R2, which leads to a lower overall reward.
Downstream measurement metrics for all six CeDeM, PanNuke and ISIC permutations are reported in Table~\ref{tab:cedem_permute}, Table \ref{tab:pannuke_permute} and Table ~\ref{tab:isic_permute}, while summary is present in Table~\ref{tab:permute_main}.

\begin{wraptable}{r}{0.55\textwidth}
\vspace{-1.0em}
\centering
\caption{CRM hierarchical order vs.\ primary downstream metrics. Best
\textbf{bold}, second \underline{underlined}.}
\label{tab:permute_main}
\setlength{\tabcolsep}{3pt}
\renewcommand{\arraystretch}{1.05}
\scriptsize
\begin{tabular}{@{}lcccccc@{}}
\toprule
& \multicolumn{2}{c}{\textbf{CeDeM}} & \multicolumn{2}{c}{\textbf{PanNuke}} & \multicolumn{2}{c}{\textbf{ISIC}} \\
\cmidrule(lr){2-3}\cmidrule(lr){4-5}\cmidrule(lr){6-7}
\textbf{Order} & MAE$_\text{R}\!\downarrow$ & MRE$_\text{R}\!\downarrow$ & mDice$\uparrow$ & mIoU$\uparrow$ & Acc.$\uparrow$ & F1$\uparrow$ \\
\midrule
$\boldsymbol{R_1R_2R_3}$ & \textbf{39.58} & \textbf{17.24} & \textbf{61.54} & \textbf{45.77} & 64.00 & \textbf{50.00} \\
$R_1R_3R_2$              & \underline{40.49} & 18.54 & \underline{59.87} & \underline{44.06} & 62.76 & 46.57 \\
$R_2R_1R_3$              & 44.37 & 20.95 & 59.52 & 43.60 & 59.06 & 45.99 \\
$R_2R_3R_1$              & 41.13 & \underline{18.71} & 59.75 & 43.71 & \underline{64.84} & 47.38 \\
$R_3R_1R_2$              & 45.83 & 21.04 & 57.97 & 42.73 & \textbf{65.07} & \underline{47.67} \\
$R_3R_2R_1$              & 44.83 & 20.14 & 58.91 & 43.38 & 64.07 & 45.16 \\
\bottomrule
\end{tabular}
\vspace{-1em}
\end{wraptable}


\paragraph{Limitations.}
\label{sec:limitations}

\ours{} requires domain expertise in designing task specific sub rewards, including choosing verifiers, and defining the hierarchical order. 
However, in cases of images with class level distinct attributes, improving the fine grain appearance cues such as intensity and texture, before a final task reward, is beneficial as observed on PanNuke, and ISIC.


\section{Conclusion}

\label{sec:conclusion}


We presented \ours{}, a framework for aligning mask conditional diffusion models using a Compositional Reward Model (CRM) with hierarchical constrained propagation (HCP). Instead of a single scalar reward, CRM decomposes image quality into naturally occurring verifier grounded sub rewards targeting distinct failure modes, including intensity, texture, structural consistency, and semantic correctness. HCP enforces a fine to coarse reward hierarchy, preventing higher level objectives from masking low level errors. Combined with GDPO, \ours{} consistently improves over SFT, ORM, and the SUM/MaxMin aggregation baselines across three downstream task types across PanNuke, CeDeM, and ISIC. Same recipe can be used for other conditional medical image generation applications. We will release code, models and generated dataset for further study.


\bibliographystyle{plainnat}
\bibliography{references}

\newpage
\appendix

\section{Additional Experimental Details}
\label{app:details}

\subsection{Dataset Visualization}
\label{app:dataset_examples}

\begin{figure}[h]
\centering
\begin{tikzpicture}[
    picture format/.style={inner sep=1pt},
]


\node[picture format] (A1)
    {\includegraphics[width=1in,height=1in]{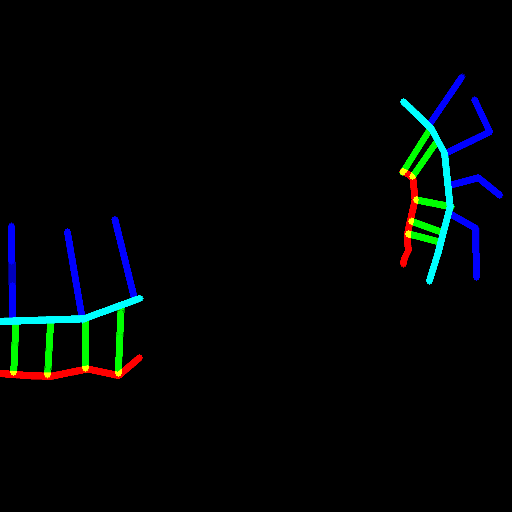}};

\node[picture format, anchor=north west] (A2) at (A1.north east)
    {\includegraphics[width=1in,height=1in]{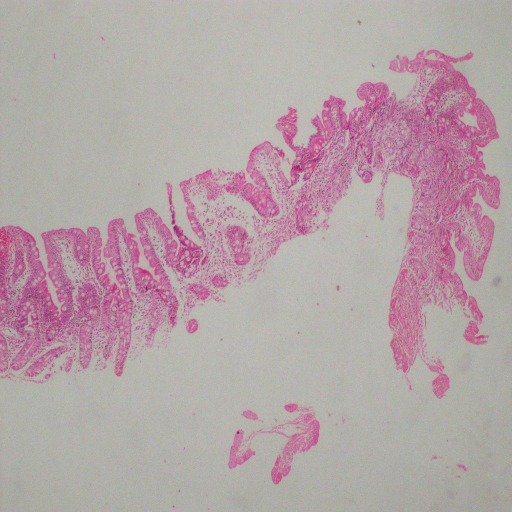}};

\node[picture format, anchor=north west] (A3) at (A2.north east)
    {\includegraphics[width=1in,height=1in]{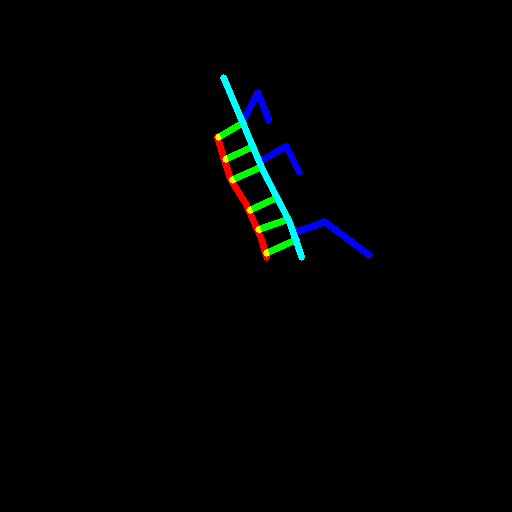}};

\node[picture format, anchor=north west] (A4) at (A3.north east)
    {\includegraphics[width=1in,height=1in]{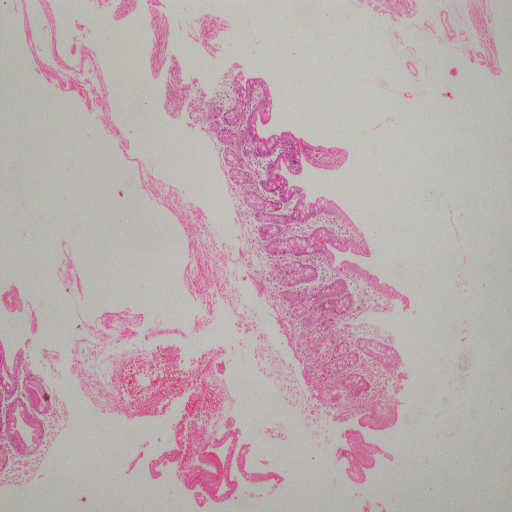}};


\node[picture format, anchor=north] (B1) at (A1.south)
    {\includegraphics[width=1in,height=1in]{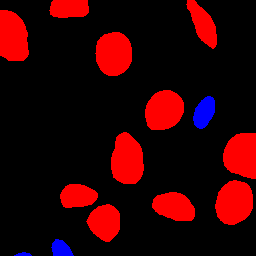}};

\node[picture format, anchor=north] (B2) at (A2.south)
    {\includegraphics[width=1in,height=1in]{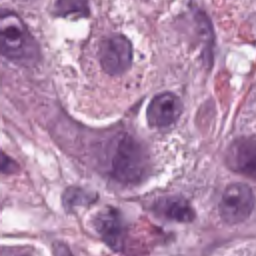}};

\node[picture format, anchor=north] (B3) at (A3.south)
    {\includegraphics[width=1in,height=1in]{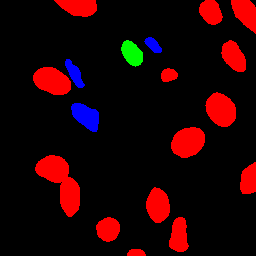}};

\node[picture format, anchor=north] (B4) at (A4.south)
    {\includegraphics[width=1in,height=1in]{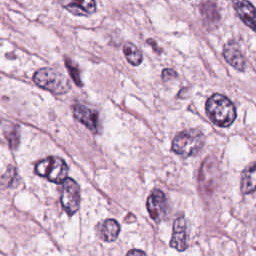}};

\node[picture format, anchor=north] (C1) at (B1.south)
    {\includegraphics[width=1in,height=1in]{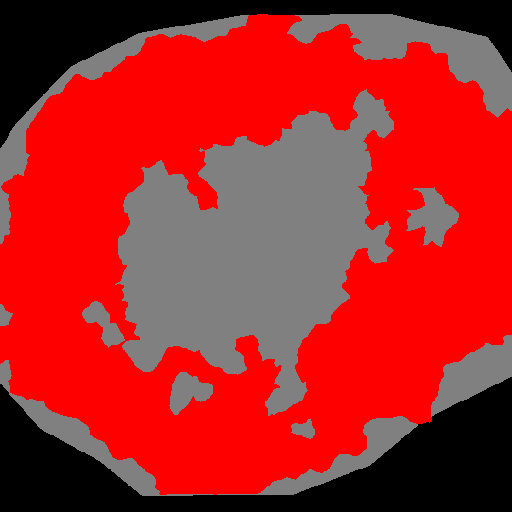}};

\node[picture format, anchor=north] (C2) at (B2.south)
    {\includegraphics[width=1in,height=1in]{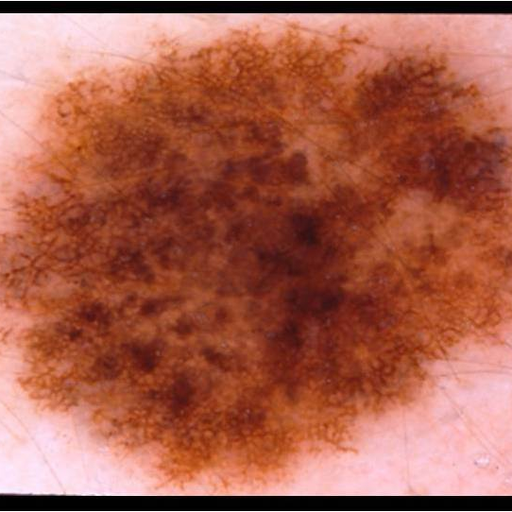}};

\node[picture format, anchor=north] (C3) at (B3.south)
    {\includegraphics[width=1in,height=1in]{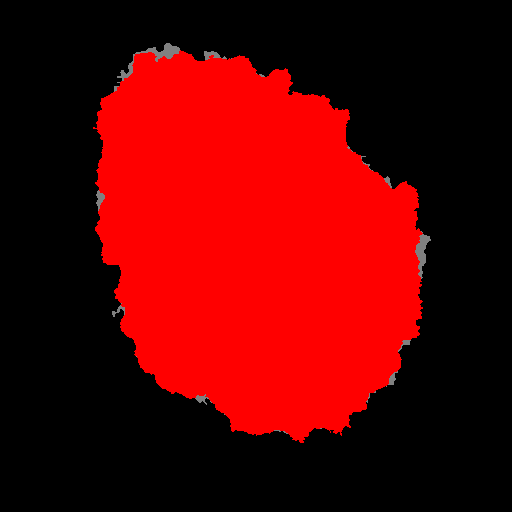}};

\node[picture format, anchor=north] (C4) at (B4.south)
    {\includegraphics[width=1in,height=1in]{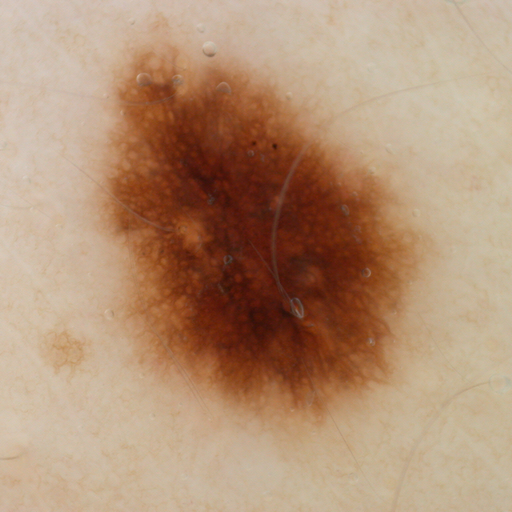}};


\node[anchor=south] at (A1.north) {\scriptsize \textbf{Mask}};
\node[anchor=south] at (A2.north) {\scriptsize \textbf{Image}};
\node[anchor=south] at (A3.north) {\scriptsize \textbf{Mask}};
\node[anchor=south] at (A4.north) {\scriptsize \textbf{Image}};


\node[rotate=90, anchor=south] at (A1.west) {\scriptsize \textbf{CeDeM}};
\node[rotate=90, anchor=south] at (B1.west) {\scriptsize \textbf{PanNuke}};
\node[rotate=90, anchor=south] at (C1.west) {\scriptsize \textbf{ISIC}};

\end{tikzpicture}

\caption{Dataset examples used for mask-conditional generation.
Each row shows two representative conditioning masks and their
paired real target images. CeDeM masks encode villus (blue), crypt (green), villi shoudler (teal) and crypt border (red) for measurement, PanNuke masks encode nuclei classes for
segmentation, and ISIC masks encode lesion extent and dermoscopic
attributes for diagnostic generation.}
\label{fig:dataset_examples}
\end{figure}
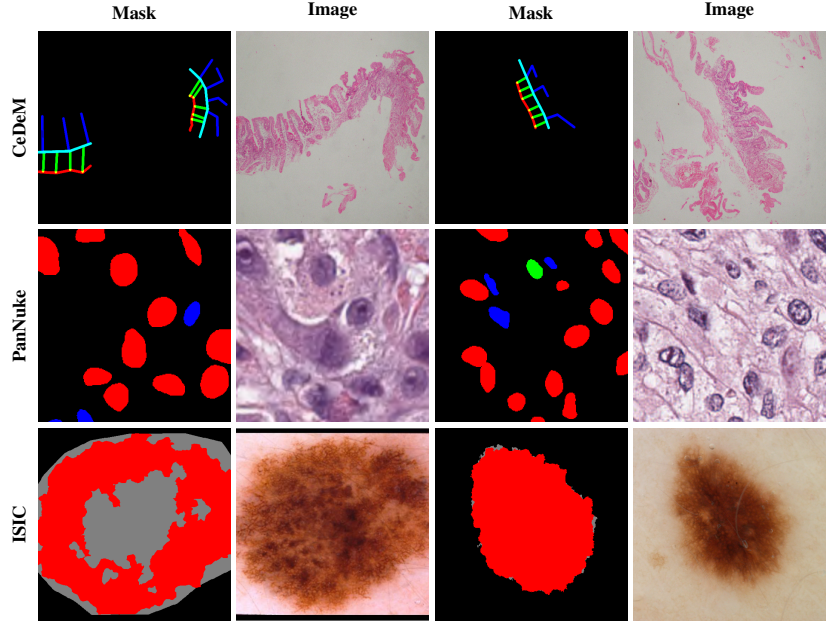

Figure~\ref{fig:dataset_examples} shows representative conditioning
mask--target pairs for the three benchmarks. The masks are the inputs
given to ControlNet; the paired images are the real targets used during
supervised training. CeDeM masks are thickened only for display.

\subsection{ControlNet Architecture and Training}
\label{app:controlnet}

We fine-tune ControlNet~\cite{zhang2023adding} on Stable Diffusion
2.1~\cite{rombach2022high} separately for each dataset. A semantic
mask is converted to a dataset-specific colour map and paired with a
text prompt. The image is encoded as latent $\mathbf{z}_0$; after
adding Gaussian noise at timestep $t$, the ControlNet branch predicts
the noise conditioned on the latent, mask, prompt, and timestep. The
supervised loss is
\begin{equation}
    \mathcal{L}_\mathrm{sup}(\theta)
    = \mathbb{E}_{(I,M,T),t,\epsilon}
    \left\|\epsilon_\theta(\mathbf{z}_t,\mathbf{c}(M),T,t)-\epsilon\right\|_2^2.
    \label{eq:supervised_denoising_loss}
\end{equation}

\begin{wrapfigure}{r}{0.48\linewidth}
\centering
\vspace{-2.0em}
\includegraphics[width=\linewidth]{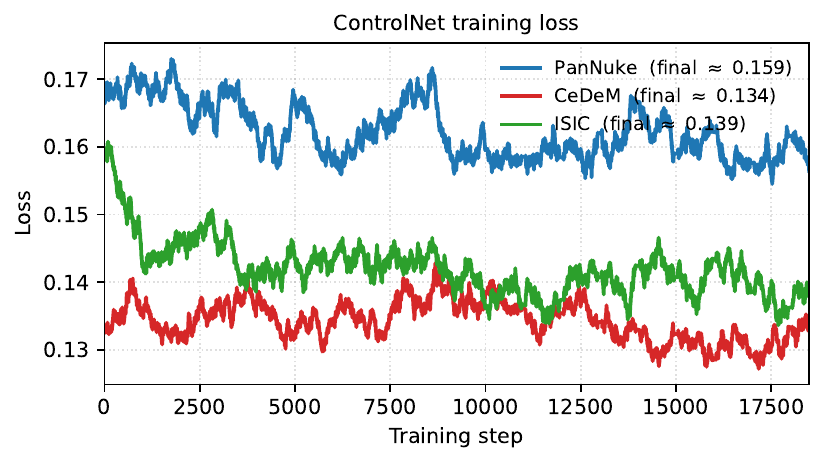}
\caption{Supervised ControlNet training loss for the three datasets. Curves are EMA-smoothed over the raw per-step diffusion loss}
\label{fig:controlnet_loss}
\vspace{-2em}
\end{wrapfigure}
Only the ControlNet branch is updated; the pretrained diffusion
backbone and text encoder remain frozen. The selected supervised
checkpoint defines $\pi_0$ for all GDPO runs.

\paragraph{Training curves.}

Figure~\ref{fig:controlnet_loss} plots EMA-smoothed supervised losses
over the shared step range. The smoothing removes timestep noise from
the diffusion objective and shows that all three initial policies reach
a stable plateau before GDPO fine-tuning.

\paragraph{Best-checkpoint selection.}
We select checkpoints by validation FID rather than raw diffusion
loss. Every $E$ epochs, a fixed validation i.e. 1000 prompt/mask pool is sampled
and compared with the corresponding real validation images; the lowest
FID checkpoint is used as $\pi_0$. The same criterion selects the best
checkpoint within each GDPO run.

\paragraph{Supervised ControlNet hyperparameters.}
All supervised runs use AdamW with learning rate $10^{-5}$, batch
size $4$ per GPU, $512{\times}512$ resolution, and up to $250$
epochs; only the ControlNet branch and zero-conv injection layers are
trainable. 

\subsubsection{PanNuke CRM}
\label{app:crm_pannuke}

PanNuke uses the ordered reward tuple
$(r_\mathrm{int}, r_\mathrm{tex}, r_\mathrm{seg})$
defined in Eq.~\ref{eq:pannuke_rewards}. The input mask has one
background class and five foreground nuclei classes. We evaluate
the first three rewards per foreground class and average over
$\mathcal{C}_f$ so that rare classes contribute equally to common
ones.

\paragraph{Intensity.}
The deployed intensity verifier uses a precomputed reference bank
instead of fitting statistics online during reward evaluation. For
each foreground class $c$, we collect nuclei pixels from real
PanNuke images using the ground-truth masks and store Lab-space
summaries in a class-conditional bank $\mathcal{B}_c$. Each bank
entry contains the per-channel mean, variance, and low-resolution
histogram of a real nucleus instance. A class-conditional GMM $P_c$
is fit once to this bank. At reward time, the conditioning mask
selects generated pixels $\mathcal{X}_c$ for class $c$, the same Lab
summary is computed, and $R_\mathrm{GMM}(\mathcal{X}_c;P_c)$ scores
whether the generated class appearance lies on the real intensity
manifold. Empty class regions are skipped in the class average.

\paragraph{Texture.}
For each class region, we extract Gray-Level Co-occurrence Matrix
statistics and multi-scale Gabor responses, concatenate them into
$\mathcal{T}_c$, and score them with a class-conditional descriptor
GMM $Q_c$. This stage penalizes samples whose color statistics are
reasonable but whose nuclear texture is over-smoothed or repetitive.

\paragraph{Segmentation.}
A frozen SegFormer-B3 trained only on real PanNuke predicts
$\phi(\mathbf{x}_0)$. The segmentation reward is macro-Dice against
the conditioning mask. This final stage checks whether the generated
image still supports the semantic label map that it was conditioned
on.

\subsubsection{CeDeM CRM}
\label{app:crm_cedem}

CeDeM uses the ordered reward tuple
$(r_\mathrm{anchor},r_\mathrm{morph},r_\mathrm{ratio})$, matching
Section~\ref{subsec:sub_rewards}. The verifier first runs the frozen
segmenter followed by MeasureNet on $\mathbf{x}_0$ to obtain predicted
villus and crypt masks, shoulder anchors, crypt borders, and the
derived villus-to-crypt ratio. Rewards are then computed from these
predicted geometric primitives against the conditioning mask and its
annotations.

\paragraph{Anchor consistency.}
This component corresponds to $r_\mathrm{anchor}$ in
Eq.~\ref{eq:cedem_Anchor}. It checks the local anatomical anchors that
drive the measurement: villus shoulders (VS) and the adjacent crypt
border (CB). The segmentation term $d_\mathrm{seg}$ measures agreement
between the predicted and conditioning villus/crypt support near these
anchors. The contour term $d_\mathrm{dtw}$ aligns CB each VS with Dynamic Time Warping. This reward is therefore an anchor-and-local-contour check.

\paragraph{Morphological consistency.}
This component corresponds to $r_\mathrm{morph}$ in
Eq.~\ref{eq:cedem_morphlogy}. It compares the predicted and reference
villus/crypt polylines using Earth Mover's Distance $d_\mathrm{emd}$.
Unlike the anchor reward, which focuses on shoulder-local structure,
the morphology reward measures the global geometric agreement of the
predicted anatomical curves.

\paragraph{Ratio consistency.}
This component corresponds to $r_\mathrm{ratio}$ in
Eq.~\ref{eq:cedem_ratio}. Median villus and crypt heights are extracted
from the predicted dual mask to obtain
$\hat\rho=\hat L^\mathrm{v}/\hat L^\mathrm{c}$, which is compared with
the target ratio $\rho^\star$ using a Gaussian tolerance. This final
stage is placed after anchor and morphology checks so a sample receives
full ratio credit only when the underlying anatomical geometry is also
plausible.

\subsubsection{ISIC CRM}
\label{app:crm_isic}

ISIC uses the ordered reward tuple
$(r_\mathrm{int},r_\mathrm{attr},r_\mathrm{dx})$ defined in
Eq.~\ref{eq:isic_rewards}. The conditioning input contains a lesion
mask and diagnostic class token $y$.

\paragraph{Skin/lesion intensity.}
The conditioning mask separates lesion and surrounding skin pixels.
For each diagnostic class, a lesion-region GMM
$P_\mathrm{les}^{(y)}$ is fit on real images of that class, while a
background-skin GMM $P_\mathrm{skin}$ captures normal dermoscopic
skin appearance. The two GMM rewards are averaged so that the model
cannot satisfy the lesion color distribution by corrupting the
surrounding skin.

\paragraph{Attribute--mask consistency.}
A frozen lesion segmenter $\phi_\mathrm{les}$ predicts the lesion
extent on $\mathbf{x}_0$. Dice with the conditioning mask checks
whether the generated lesion has the requested location and shape,
independent of the diagnostic classifier.

\paragraph{Diagnostic margin.}
A frozen ResNet classifier produces logits for the diagnostic
classes. We use the margin reward in Eq.~\ref{eq:margin} rather than
raw probability because it is sensitive to the strongest competing
class and remains bounded for GDPO normalization.

\subsubsection{Gating and Normalisation}
\label{app:reward_norm}

MIN-GATE implements the prerequisite order used by the cascade. The
gate $G_k$ in Eq.~\ref{eq:cascading_reward} is computed from the
cumulative product of earlier raw component scores and lower bounded
by $\gamma=0.1$.
Thus, failing an early verifier does not make the reward exactly
zero, but it prevents later semantic or measurement rewards from
dominating the average when their prerequisites are absent. For
policy optimisation, GDPO receives the component tuple and performs
the within-group component normalization in Eq.~\ref{eq:gdpo_advantage};
the gated cascade is used as the scalar reward for logging,
selection, and aggregation comparisons.

\subsection{Mask Augmentation Pipeline}
\label{app:augmentation}

We augment conditioning masks during RL roll-outs and when constructing
synthetic downstream training sets. Empty masks are
filtered and evaluation masks are never augmented. Each source mask
yields up to ten variants.

\paragraph{Common geometric family.}
All datasets use rotations $\{90^\circ,180^\circ,270^\circ\}$ and
horizontal/vertical flips. For CeDeM, the same transforms are applied
to villus/crypt point annotations, with reflected translations of
$\{10,15\}$~px, so count and ratio prompts remain exact.

\paragraph{PanNuke (structural).}
PanNuke also uses per-instance cell dropout ($20\%$ or $30\%$ of
connected components recoloured to background) and wrap-around shifts
of $10\%$ or $15\%$ of the image side. After dropout, prompts are
regenerated from the augmented mask so class counts stay consistent.

\paragraph{ISIC.}
Lesion masks have additional structure (lesion body \texttt{+} five
dermoscopic attributes: pigment network, negative network, streaks,
milia-like cysts, globules), so we organise the augmentations into two
families that interact differently with the prompt:
\begin{itemize}[leftmargin=*,topsep=2pt,itemsep=1pt]
    \item \textbf{Geometric (prompt unchanged).}
    Rotations $\{90^\circ,180^\circ,270^\circ\}$, horizontal/vertical flips,
    no-wrap shifts of $\{10\%,15\%\}$ (vacated region filled with background
    rather than rolled), and zoom-in $1.25\times$ (centre-crop and resize) /
    zoom-out $0.80\times$ (centre-pad and resize). All use nearest-neighbour
    interpolation to preserve attribute colours.
    \item \textbf{Semantic (mask + prompt both change).} For each attribute
    $a\!\in\!\{\text{pigment network, negative network, streaks, milia-like
    cysts, globules}\}$, \texttt{drop\_attr\_$a$} demotes that attribute's
    pixels to lesion-body colour and removes the corresponding token from
    the ``\dots showing \dots'' clause; \texttt{drop\_all\_attrs} collapses
    every attribute to body for a clean ablation condition. Variants whose
    target attribute is absent in the source mask are skipped.
\end{itemize}

\subsection{GDPO Hyperparameters}
\label{app:gdpo}
We adopt the GDPO update of~\citet{liu2026gdpo} with PPO-style
clipping (Eq.~\ref{eq:gdpo_loss}); the rollout policy is
refreshed every inner-iteration block, and per-component
advantages use a per-prompt running buffer. Common settings are
listed in Table~\ref{tab:gdpo_hparams}.

\begin{table}[h]
\centering
\caption{GDPO hyperparameters per dataset. ``Group $N$'' is the
number of candidate samples per conditioning prompt; advantages
are computed per component within each group and renormalised
before clipping (Eq.~\ref{eq:gdpo_advantage}). ``Inner iters'' is
the number of policy-update passes per rollout batch.}
\label{tab:gdpo_hparams}
\setlength{\tabcolsep}{4pt}
\small
\begin{tabular}{@{}lccc@{}}
\toprule
& \textbf{CeDeM} & \textbf{PanNuke} & \textbf{ISIC} \\
\midrule
Optimizer                       & AdamW & AdamW & AdamW \\
Learning rate                   & $5{\times}10^{-6}$ & $5{\times}10^{-6}$ & $5{\times}10^{-6}$ \\
Batch size (per GPU)            & 4 & 4 & 4 \\
Group size $N$                  & 32 & 32 & 32 \\
Clip ratio $\epsilon_\mathrm{clip}$ & 0.1 & 0.1 & 0.1 \\
Inner iters                     & 2 & 2 & 2 \\
DDIM sampling steps             & 100 & 100 & 100 \\
Rollout buffer size             & 64 & 128 & 128 \\
Min prompt count $m$            & 8 & 8 & 8 \\
Total RL epochs                 & 100 & 200 & 200 \\
Min gate $\gamma$               & 0.1 & 0.1 & 0.1 \\
\bottomrule
\end{tabular}
\end{table}

\subsection{Downstream Model Training}
\label{sec:downstream}

\subsubsection{SegFormer-B3 (PanNuke)}
\label{sec:supp_downstream_segformer_pannuke}
We use the \texttt{nvidia/mit-b3} encoder pretrained on
ImageNet-1k with a 6-class semantic head (background + five nuclei
classes). Inputs are resized to $256{\times}256$ and standardised
with ImageNet mean/std. The training objective is a sum of
pixel-wise cross-entropy and soft-Dice. We optimise with AdamW
(learning rate $6{\times}10^{-5}$, weight decay $0.01$), a
polynomial learning-rate schedule (power 1.0) over 250 epochs,
and a batch size of 16; evaluation is performed every epoch on the
real validation split and the best mean-IoU checkpoint is kept.

\subsubsection{ResNet-50 (ISIC)}
The ISIC diagnostic classifier is a torchvision ResNet-50 with
ImageNet-V2 weights, the final linear layer replaced by a 3-way
head (nevus / melanoma / seborrheic keratosis). Each input is the
lesion crop extracted from the conditioning lesion mask, padded by
$10\%$ and resized to $224{\times}224$. We train for 200 epochs
with AdamW (learning rate $3{\times}10^{-4}$, weight decay
$10^{-4}$), cosine annealing to zero, and a batch size of 32. 

\subsubsection{MeasureNet (CeDeM)}
\label{sec:supp_downstream_measurenet}

CeDeM annotations mark villi and crypts by their bisector polylines,
with additional labels for villus shoulders and crypt borders. We use
the detection part of MeasureNet~\cite{tyagi2024measurenet} as the
downstream model. MeasureNet is built as a polyline detector: it
predicts villus and crypt bisectors, from which lengths and the
villus-to-crypt ratio are computed deterministically after computing the polyline length. We optimizer Adam, learning rate $10^{-4}$, batch size 8, $\sim$200 epochs. Additionally, we train a segmentation model (segformer-b5 \citep{xie2021segformer}) for villi shoulder and crypt border. Same segmentation model is used as verifier in CRM (Appendix~\ref{app:crm_cedem}).

\section{Additional Results}
\label{app:results}

\subsection{Full Per-Dataset Results}
\label{app:full_metrics}

This section reports the complete per dataset metric tables summarized in Table~\ref{tab:main_results} of the main paper. Table~\ref{tab:cedem_full} provides additional measurement metrics for CeDeM, including Villi and Crypt MAE$_V$, MRE$_V$, MAE$_C$, and MRE$_C$. Table~\ref{tab:isic_full} reports the complete ISIC classification metrics.

\begin{table}[h]
\centering
\caption{CeDeM full measurement metrics. Downstream model:
MeasureNet~\cite{tyagi2024measurenet}. Subscripts denote
villi (V), crypt (C), and villus-to-crypt ratio (R). Lower is
better for all columns.}
\label{tab:cedem_full}
\resizebox{\textwidth}{!}{%
\begin{tabular}{@{}llcccccc@{}}
\toprule
\textbf{Method} & \textbf{Reward} & MAE$_\text{V}\downarrow$ & MRE$_\text{V}\downarrow$ & MAE$_\text{C}\downarrow$ & MRE$_\text{C}\downarrow$ & MAE$_\text{R}\downarrow$ & MRE$_\text{R}\downarrow$ \\
\midrule
SFT (Original only)              & --                                & 13.88          & 14.56          & 8.41          & 21.52          & 52.06          & 24.77 \\
Generated (SFT)                  & --                                & 11.88          & 11.74          & 7.87          & 17.84          & 47.69          & 22.15 \\
Best-of-N                        & Inference                         & 10.26          & 10.49          & 7.55          & 17.84          & 45.71          & 21.83 \\
\midrule
GDPO~\cite{liu2026gdpo}          & ORM (Ratio)                       & 11.39          & 10.98          & 7.02          & 15.67          & 41.95          & 18.85 \\
GDPO~\cite{liu2026gdpo}          & SUM~\cite{liu2026gdpo}            & 9.89           & 10.55          & 7.18          & 16.58          & 45.33          & 21.41 \\
GDPO~\cite{liu2026gdpo}          & MaxMin                            & 10.76          & 10.26          & 7.35          & 16.44          & 47.07          & 21.55 \\
\textbf{GDPO (ours)}             & \textbf{CRM}                      & \underline{10.01} & \underline{9.68} & \textbf{6.64} & \textbf{14.32} & \textbf{39.58} & \textbf{17.24} \\
\bottomrule
\end{tabular}

}
\end{table}

\begin{table}[h]
\centering
\caption{ISIC full skin-lesion classification metrics.
Downstream model: ResNet-50 trained on real$+$synthetic
dermoscopy images and evaluated on the held-out real
ISIC-2019 test split (3 classes: nevus, melanoma, seborrheic
keratosis). Precision, recall, and F1 are macro-averaged over
the three classes. Higher is better.}
\label{tab:isic_full}
\begin{tabular}{@{}llcccc@{}}
\toprule
\textbf{Method} & \textbf{Reward} & Acc.$\uparrow$ & Prec.$\uparrow$ & Rec.$\uparrow$ & F1$\uparrow$ \\
\midrule
SFT (Original only)              & --                                & 53.43          & 49.91          & 50.55          & 44.84 \\
Generated (SFT)                  & --                                & 55.82          & \underline{50.93} & 53.16       & \underline{47.30} \\
Best-of-N                        & Inference                         & 56.00          & 49.11          & 52.74          & 46.85 \\

\midrule
GDPO~\cite{liu2026gdpo}          & ORM                               & \underline{63.00} & \textbf{54.40} & 49.90       & 45.50 \\
GDPO~\cite{liu2026gdpo}          & SUM~\cite{liu2026gdpo}            & 59.80          & 47.20          & 46.60          & 43.80 \\
GDPO~\cite{liu2026gdpo}          & MaxMin                            & 57.70          & 47.40          & 50.10          & 44.90 \\
\textbf{GDPO (ours)}             & \textbf{CRM}                      & \textbf{64.00} & \underline{51.60} & \textbf{53.60} & \textbf{50.10} \\
\bottomrule
\end{tabular}%
\end{table}

\begin{table}[h]
\centering
\caption{CeDeM cascade-order ablation with fixed components and
training budget. $R_1$: anchor, $R_2$: morphology, $R_3$: ratio;
subscripts denote villus (V), crypt (C), and ratio (R). Lower is
better. Best in \textbf{bold}, second \underline{underlined}.}
\label{tab:cedem_permute}
\begin{tabular}{@{}lcccccc@{}}
\toprule
\textbf{Order} & MAE$_\text{V}\downarrow$ & MRE$_\text{V}\downarrow$ & MAE$_\text{C}\downarrow$ & MRE$_\text{C}\downarrow$ & MAE$_\text{R}\downarrow$ & MRE$_\text{R}\downarrow$ \\
\midrule
$\boldsymbol{R_1R_2R_3}$ \textbf{(main)} & \textbf{10.01} & \textbf{9.68}  & \textbf{6.64} & \textbf{14.32} & \textbf{39.58} & \textbf{17.24} \\
$R_1R_3R_2$                              & 10.22 & 9.83              & \underline{6.50} & 15.59       & \underline{40.49} & 18.54 \\
$R_2R_1R_3$                              & 11.08 & 10.68             & 6.99          & 15.41          & 44.37          & 20.95 \\
$R_2R_3R_1$                              & 10.27 & 10.40             & 6.88          & \underline{15.12} & 41.13       & \underline{18.71} \\
$R_3R_1R_2$                              & 11.75 & 11.16             & 6.99          & 14.99          & 45.83          & 21.04 \\
$R_3R_2R_1$                              & \underline{10.09} & 10.13     & 6.96          & 15.93          & 44.83          & 20.14 \\
\bottomrule
\end{tabular}%
\end{table}

\begin{table}[h]
\centering
\caption{PanNuke cascade-order ablation with fixed components and
training budget. $R_1$: intensity, $R_2$: texture, $R_3$:
SegFormer agreement. Higher is better. Best in \textbf{bold}, second
\underline{underlined}.}
\label{tab:pannuke_permute}
\begin{tabular}{@{}lcc@{}}
\toprule
\textbf{Order} & mIoU$\uparrow$ & mDice$\uparrow$ \\
\midrule
$\boldsymbol{R_1R_2R_3}$ \textbf{(main)} & \textbf{45.77} & \textbf{61.54} \\
$R_1R_3R_2$                              & \underline{44.06} & \underline{59.87} \\
$R_2R_1R_3$                              & 43.60 & 59.52 \\
$R_2R_3R_1$                              & 43.71 & 59.75 \\
$R_3R_1R_2$                              & 42.73 & 57.97 \\
$R_3R_2R_1$                              & 43.38 & 58.91 \\
\bottomrule
\end{tabular}
\end{table}

\begin{table}[h]
\centering
\caption{ISIC cascade-order ablation with fixed components and
training budget. $R_1$: intensity, $R_2$: attribute-mask agreement,
$R_3$: diagnosis classifier. Precision, recall, and F1 are
macro-averaged; higher is better. Best in \textbf{bold}, second
\underline{underlined}.}
\label{tab:isic_permute}
\begin{tabular}{@{}lcccc@{}}
\toprule
\textbf{Order} & Acc.$\uparrow$ & Prec.$\uparrow$ & Rec.$\uparrow$ & F1$\uparrow$ \\
\midrule
$\boldsymbol{R_1R_2R_3}$ \textbf{(main)} & \underline{64.00} & \textbf{51.60} & \textbf{53.60} & \textbf{50.00} \\
$R_1R_3R_2$                              & 62.76 & 46.86 & 49.40 & 46.57 \\
$R_2R_1R_3$                              & 59.06 & 45.72 & 47.07 & 45.99 \\
$R_2R_3R_1$                              & 64.84 & \underline{49.27} & 48.76 & 47.38 \\
$R_3R_1R_2$                              & \textbf{65.07} & 47.50 & \underline{49.75} & \underline{47.67} \\
$R_3R_2R_1$                              & 64.07 & 46.42 & 45.54 & 45.16 \\
\bottomrule
\end{tabular}
\end{table}

\section{Qualitative Performance}
\label{sec:supp_qualitative}
\begin{figure}[h]
\centering
\begin{tikzpicture}[
    picture format/.style={inner sep=1pt},
]

\node[picture format] (A1)
    {\includegraphics[width=0.88in,height=0.88in]{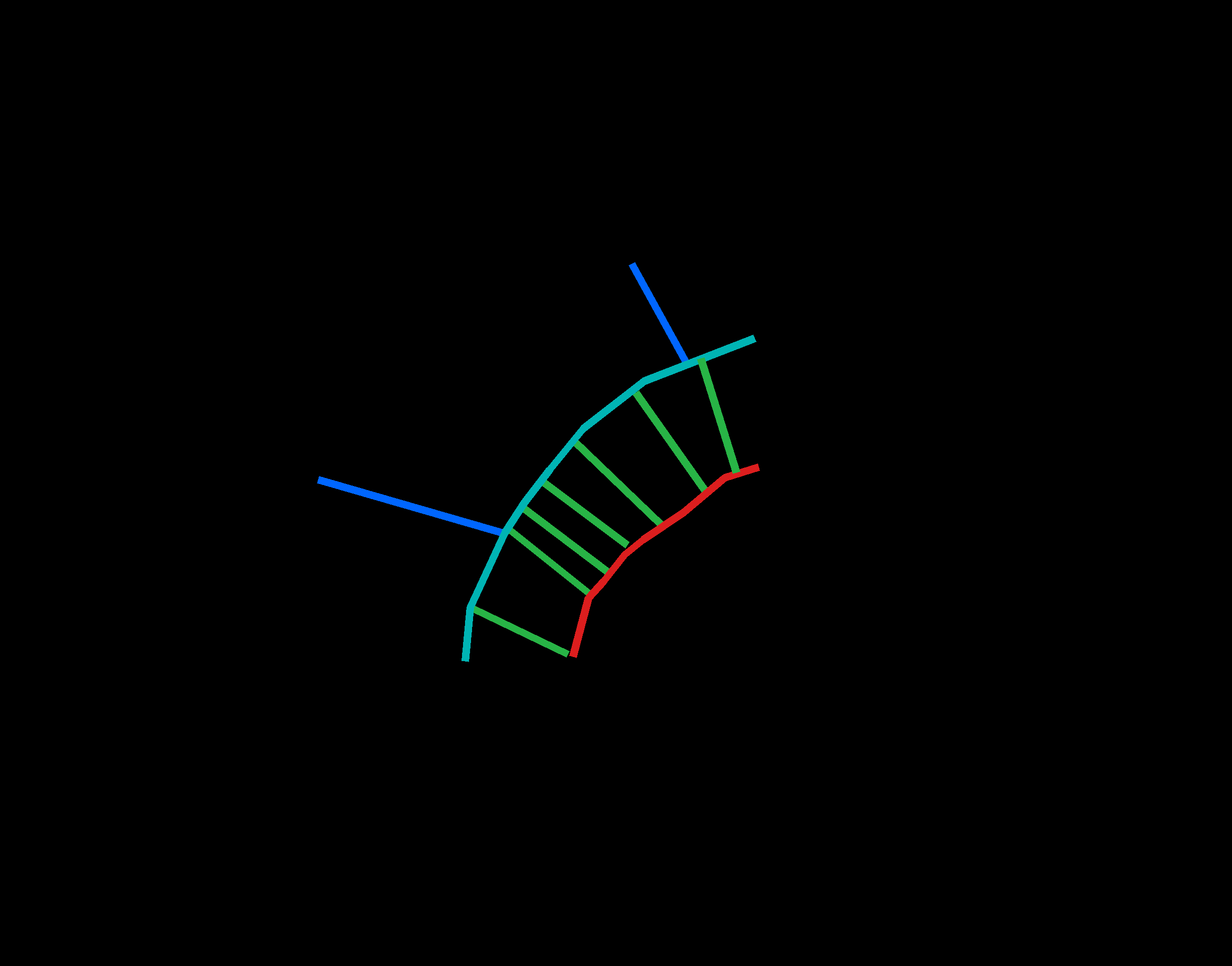}};
\node[picture format, anchor=north west] (A2) at (A1.north east)
    {\includegraphics[width=0.88in,height=0.88in]{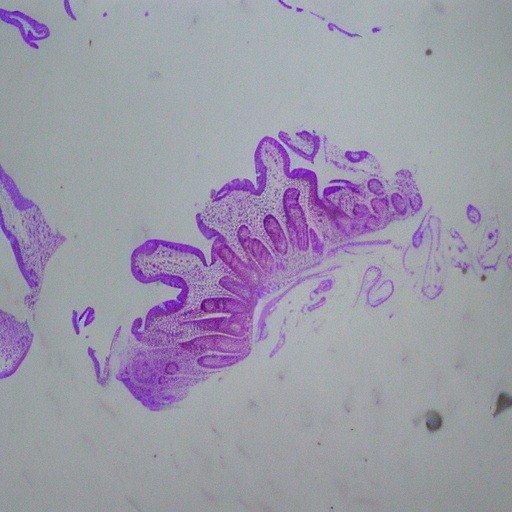}};
\node[picture format, anchor=north west] (A3) at (A2.north east)
    {\includegraphics[width=0.88in,height=0.88in]{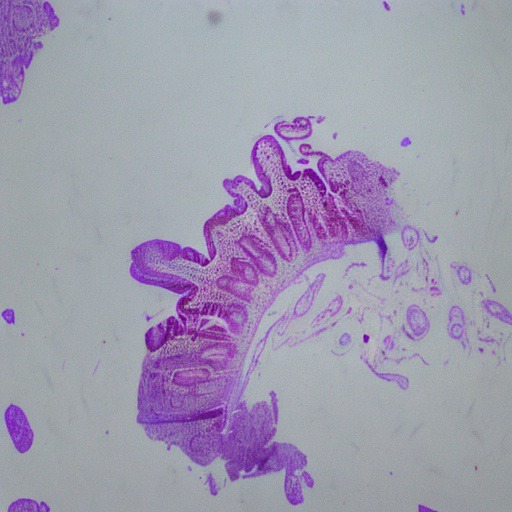}};
\node[picture format, anchor=north west] (A4) at (A3.north east)
    {\includegraphics[width=0.88in,height=0.88in]{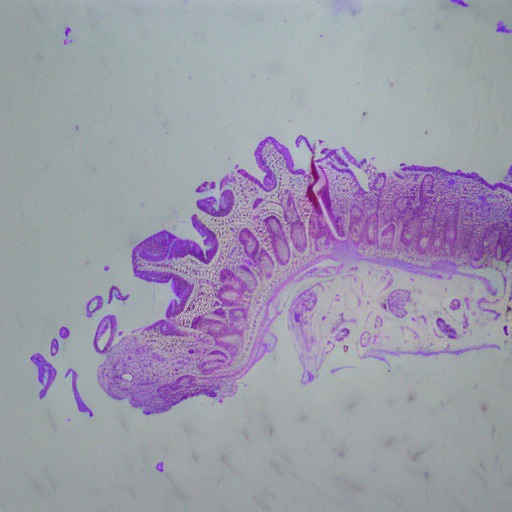}};
\node[picture format, anchor=north west] (A5) at (A4.north east)
    {\includegraphics[width=0.88in,height=0.88in]{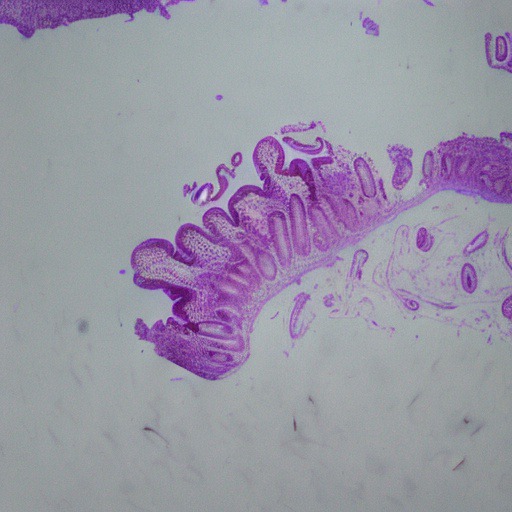}};
\node[picture format, anchor=north west] (A6) at (A5.north east)
    {\includegraphics[width=0.88in,height=0.88in]{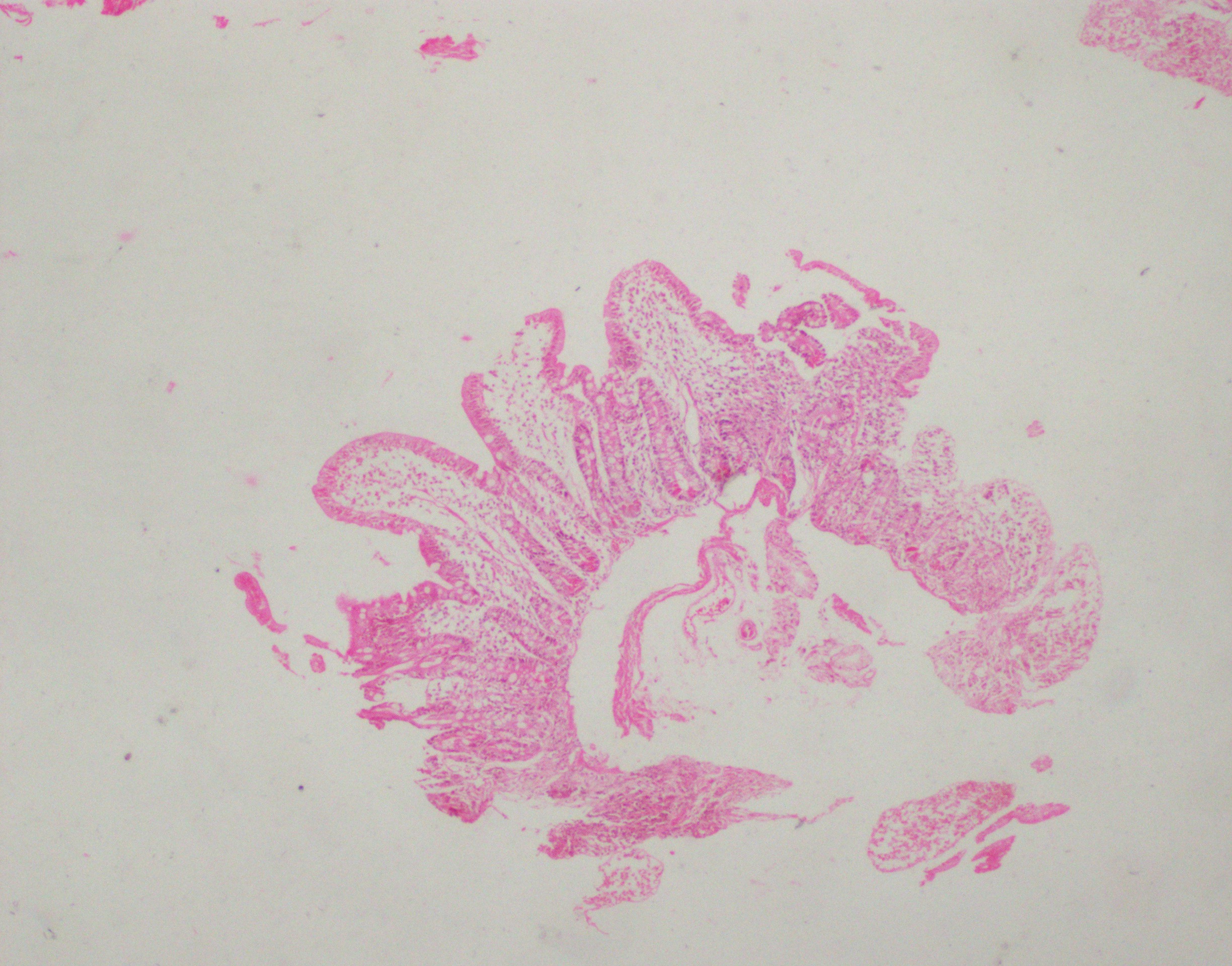}};

\node[picture format, anchor=north] (B1) at (A1.south)
    {\includegraphics[width=0.88in,height=0.88in]{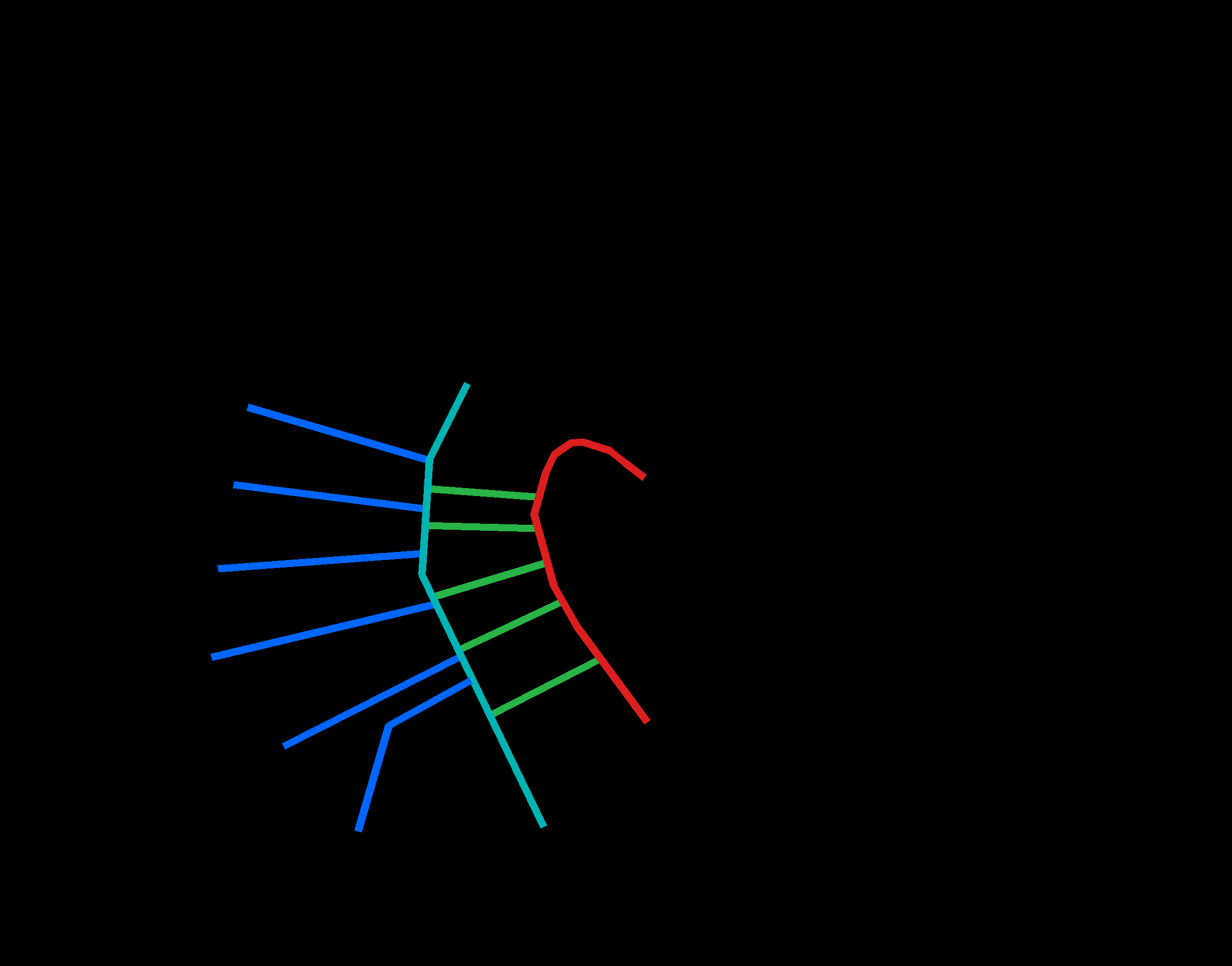}};
\node[picture format, anchor=north] (B2) at (A2.south)
    {\includegraphics[width=0.88in,height=0.88in]{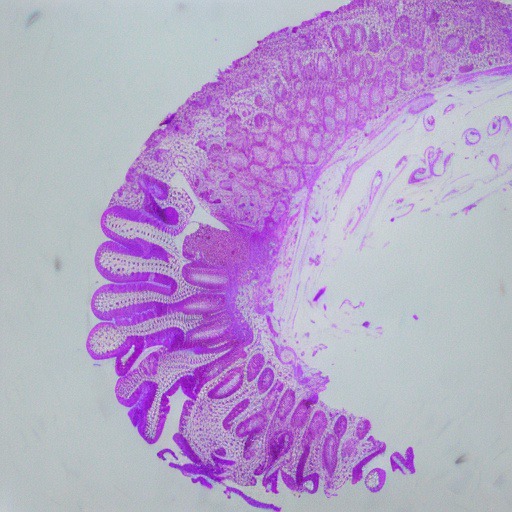}};
\node[picture format, anchor=north] (B3) at (A3.south)
    {\includegraphics[width=0.88in,height=0.88in]{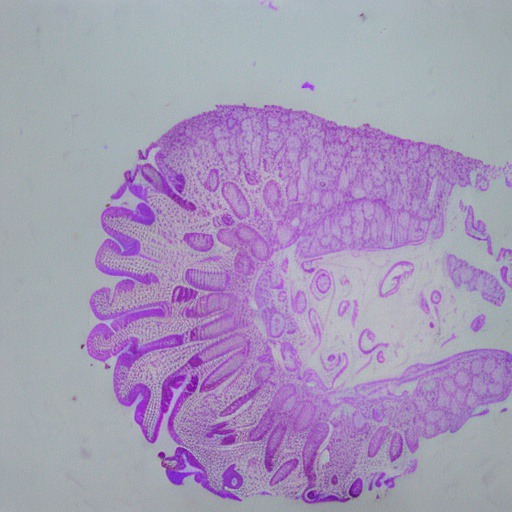}};
\node[picture format, anchor=north] (B4) at (A4.south)
    {\includegraphics[width=0.88in,height=0.88in]{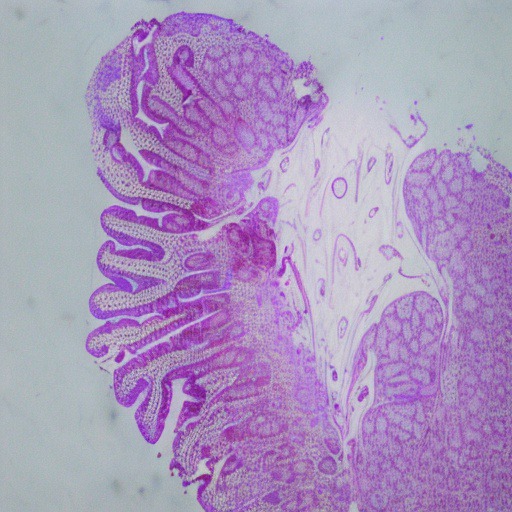}};
\node[picture format, anchor=north] (B5) at (A5.south)
    {\includegraphics[width=0.88in,height=0.88in]{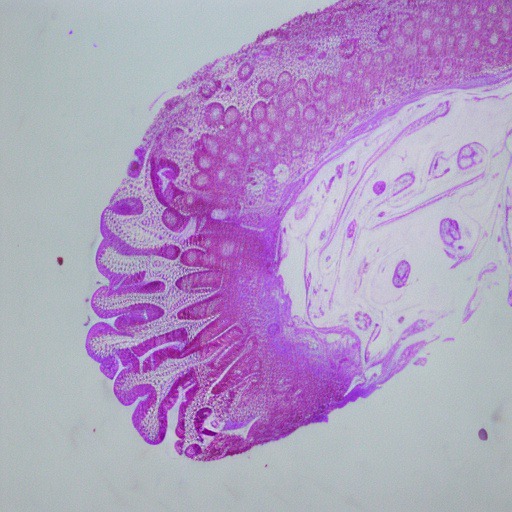}};
\node[picture format, anchor=north] (B6) at (A6.south)
    {\includegraphics[width=0.88in,height=0.88in]{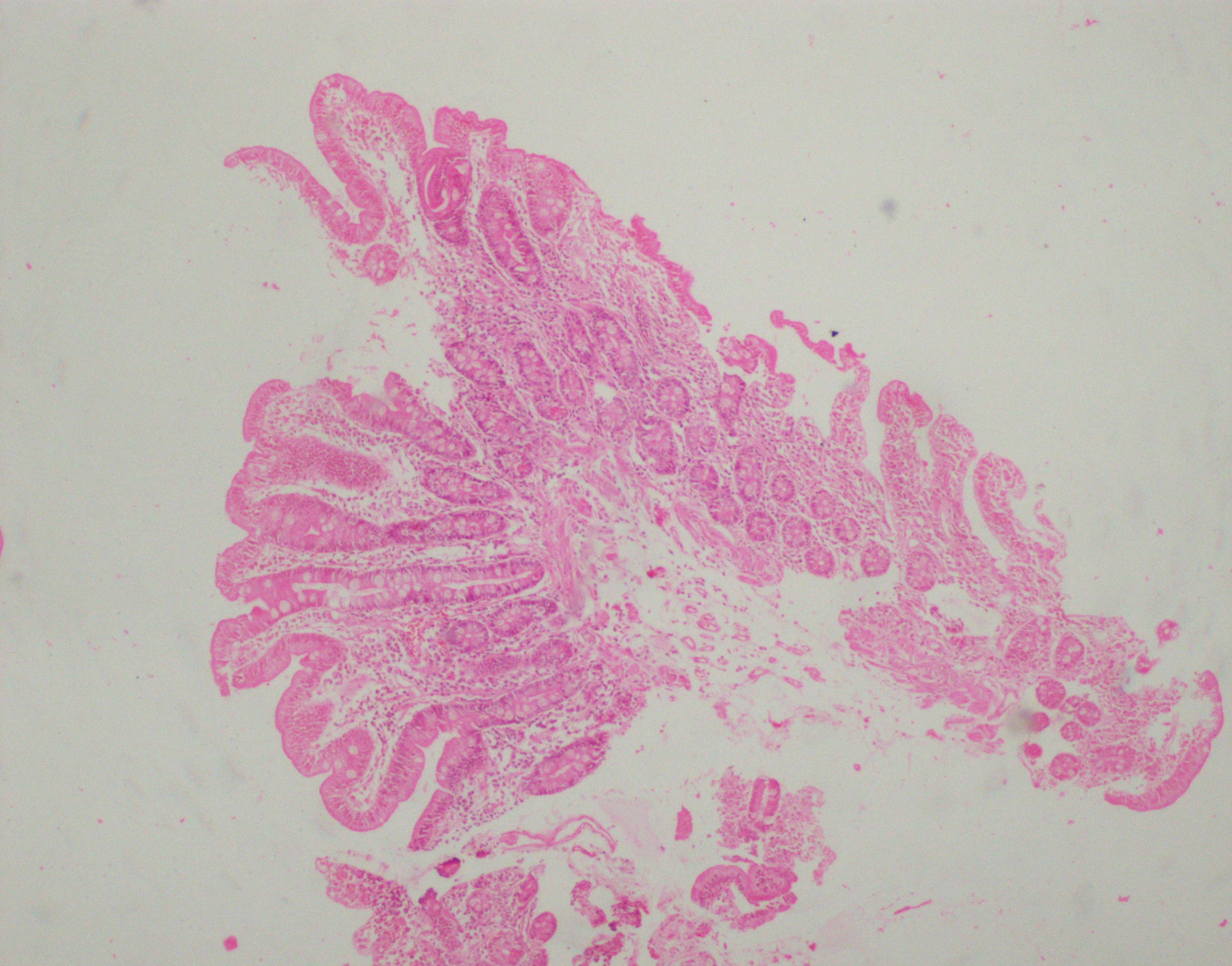}};

\node[picture format, anchor=north] (C1) at (B1.south)
    {\includegraphics[width=0.88in,height=0.88in]{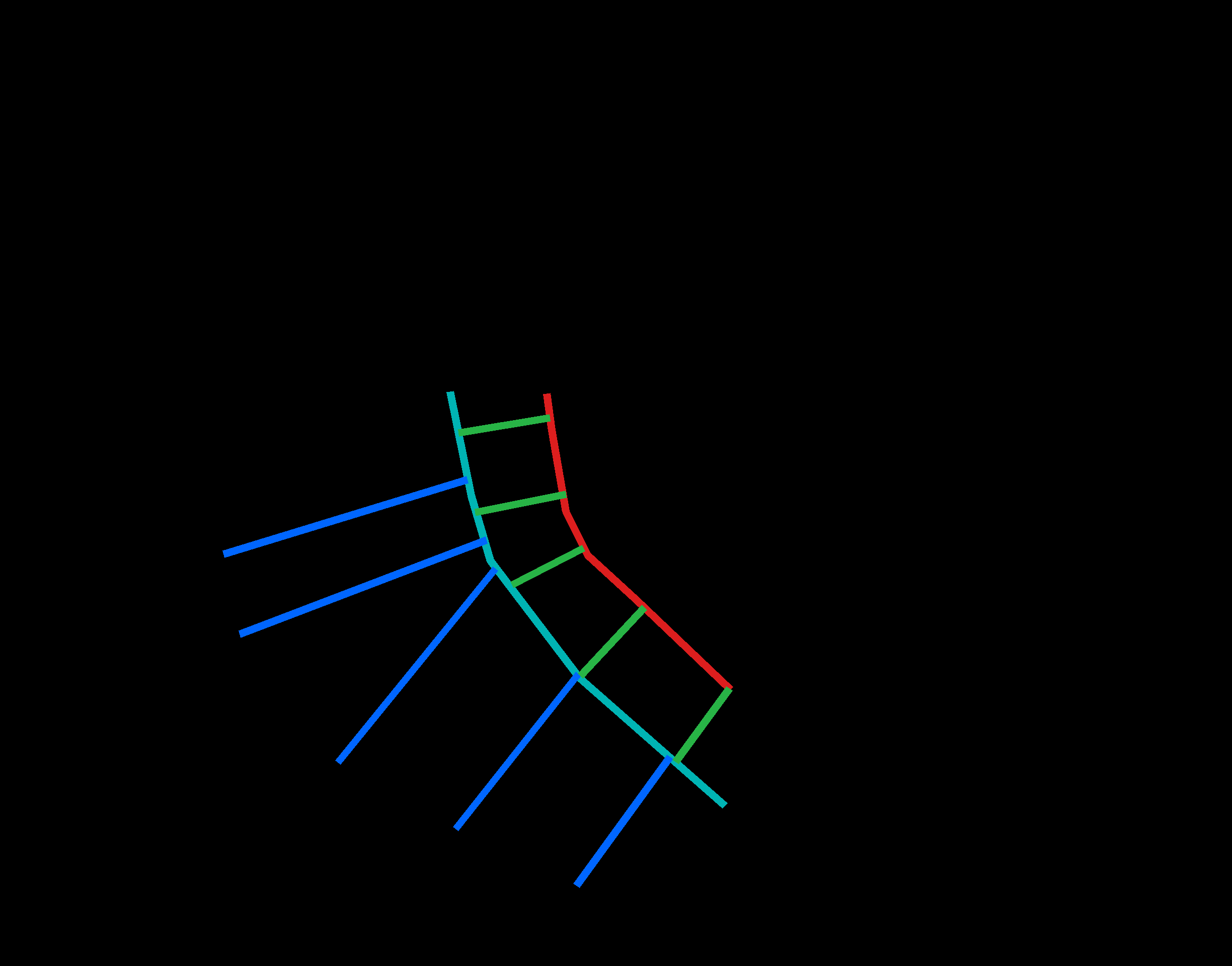}};
\node[picture format, anchor=north] (C2) at (B2.south)
    {\includegraphics[width=0.88in,height=0.88in]{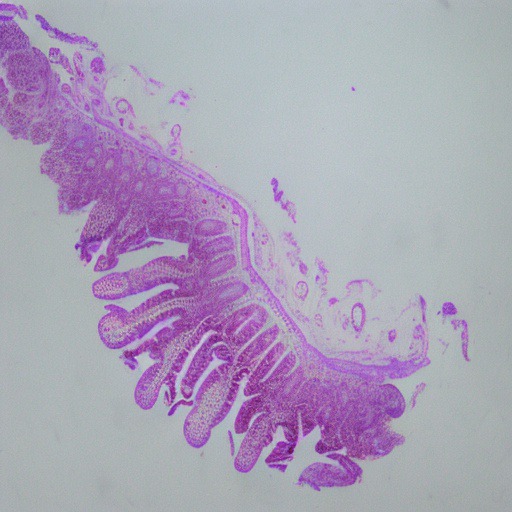}};
\node[picture format, anchor=north] (C3) at (B3.south)
    {\includegraphics[width=0.88in,height=0.88in]{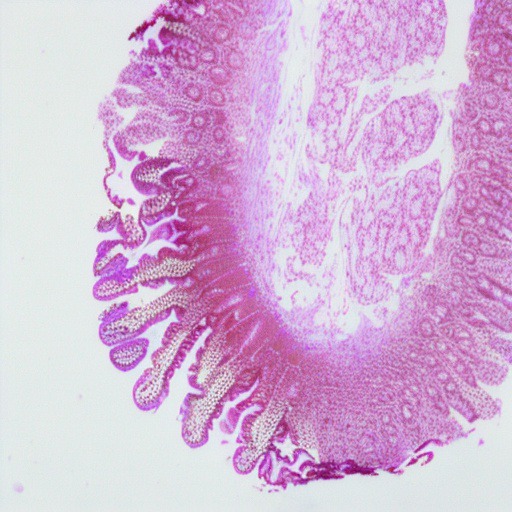}};
\node[picture format, anchor=north] (C4) at (B4.south)
    {\includegraphics[width=0.88in,height=0.88in]{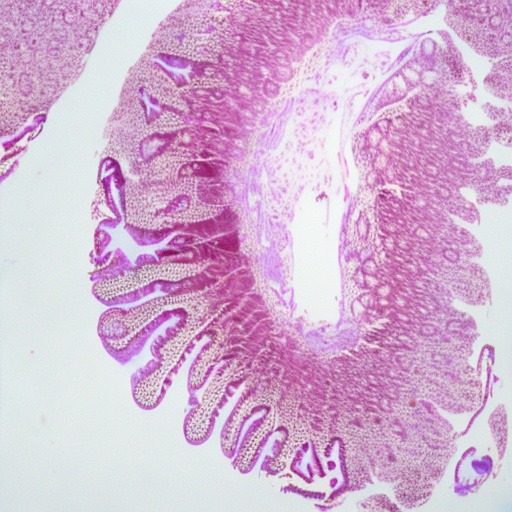}};
\node[picture format, anchor=north] (C5) at (B5.south)
    {\includegraphics[width=0.88in,height=0.88in]{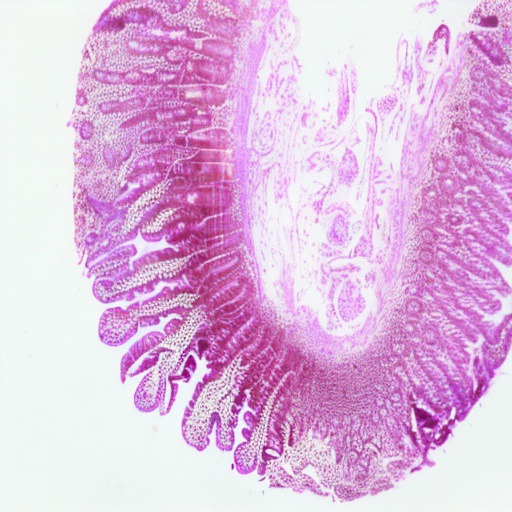}};
\node[picture format, anchor=north] (C6) at (B6.south)
    {\includegraphics[width=0.88in,height=0.88in]{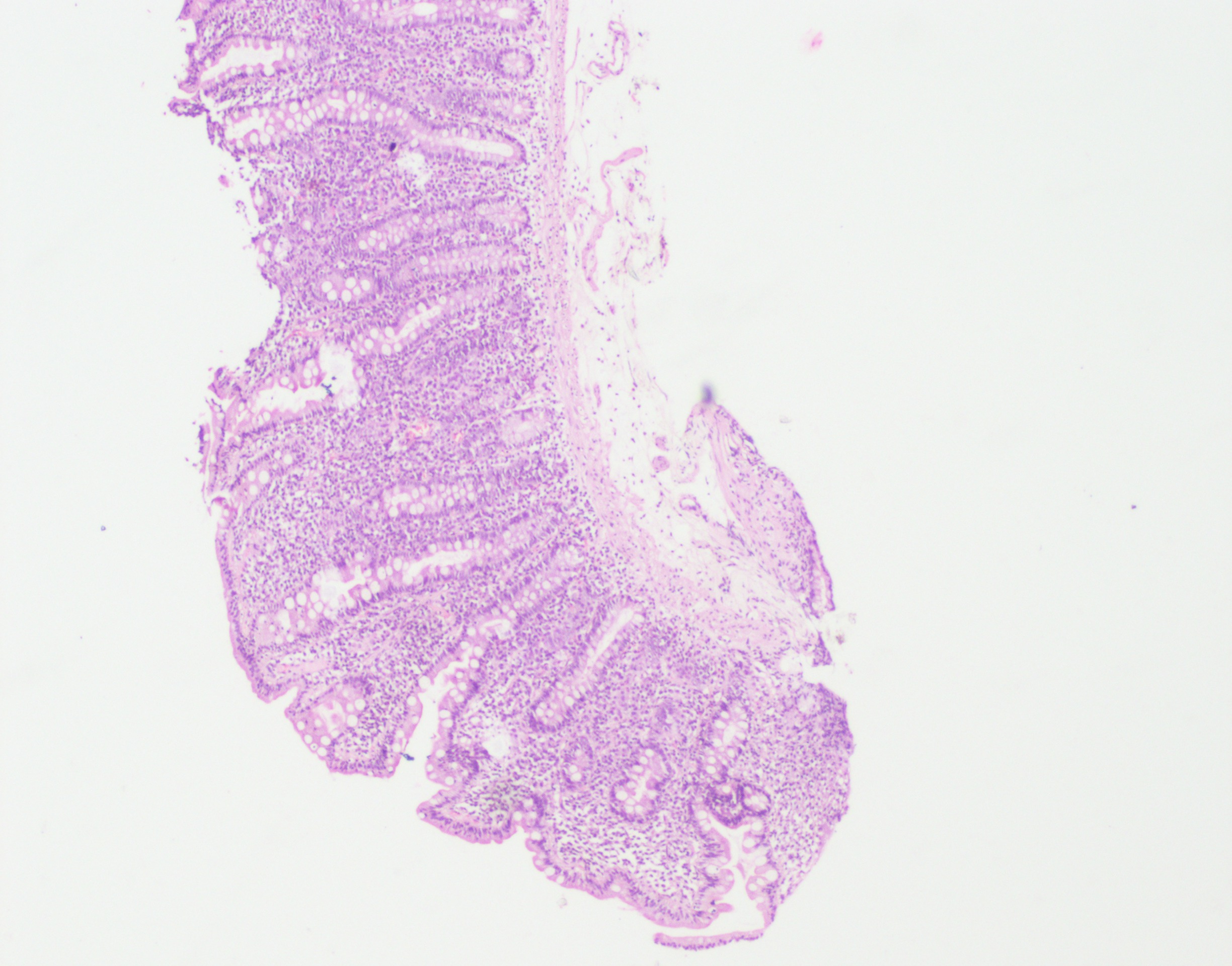}};

\node[anchor=south] at (A1.north) {\scriptsize \textbf{Mask}};
\node[anchor=south] at (A2.north) {\scriptsize \textbf{CN}};
\node[anchor=south] at (A3.north) {\scriptsize \textbf{SUM}};
\node[anchor=south] at (A4.north) {\scriptsize \textbf{MaxMin}};
\node[anchor=south] at (A5.north) {\scriptsize \textbf{CRM}};
\node[anchor=south] at (A6.north) {\scriptsize \textbf{Original}};


\end{tikzpicture}
\caption{Additional CeDeM qualitative comparisons. Mask
overlays use blue (VilliM), red (Crypt Border), green (CryptM), and
teal (Villi Shoulder). All generated columns share the conditioning
mask of the corresponding original target image.}
\label{fig:qualitative_cedem_supp}
\end{figure}
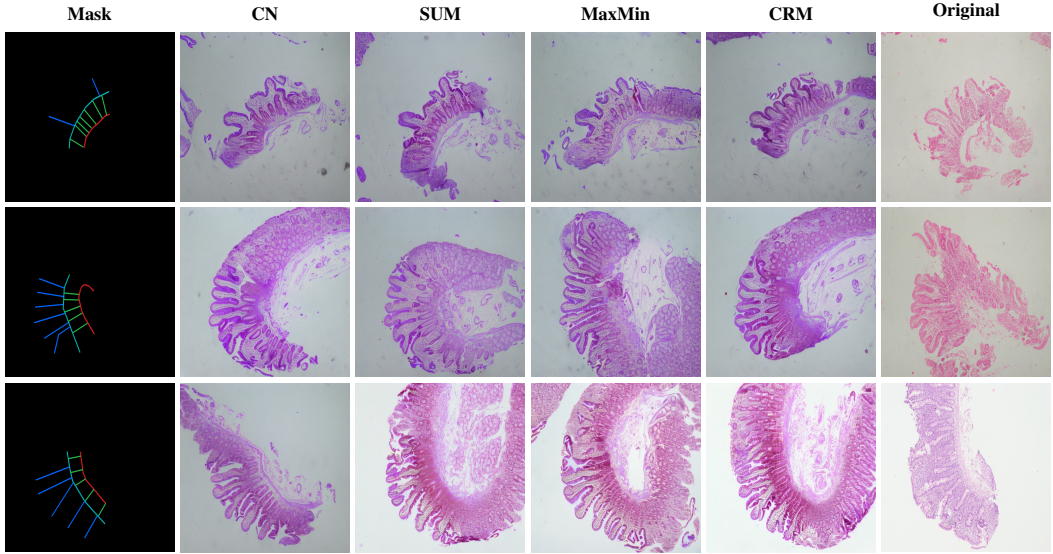

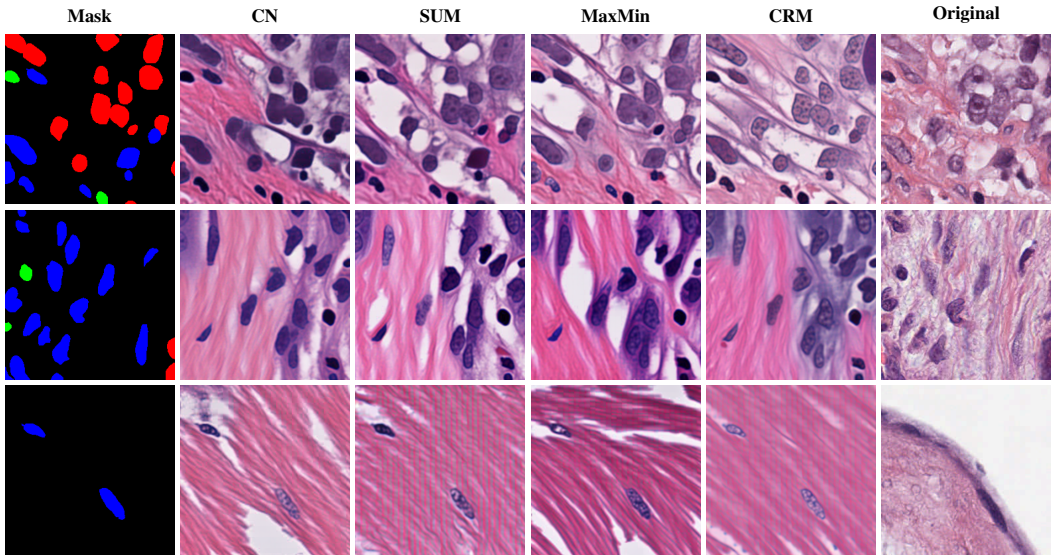
\begin{figure}[h]
\centering
\begin{tikzpicture}[
    picture format/.style={inner sep=1pt},
]

\node[picture format] (A1)
    {\includegraphics[width=0.88in,height=0.88in]{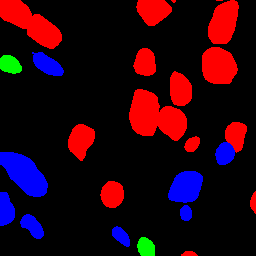}};
\node[picture format, anchor=north west] (A2) at (A1.north east)
    {\includegraphics[width=0.88in,height=0.88in]{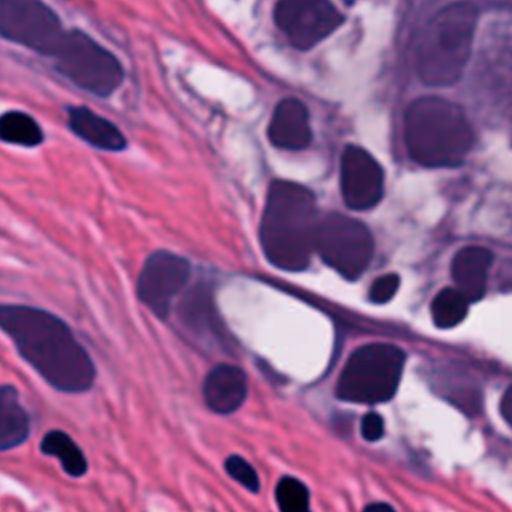}};
\node[picture format, anchor=north west] (A3) at (A2.north east)
    {\includegraphics[width=0.88in,height=0.88in]{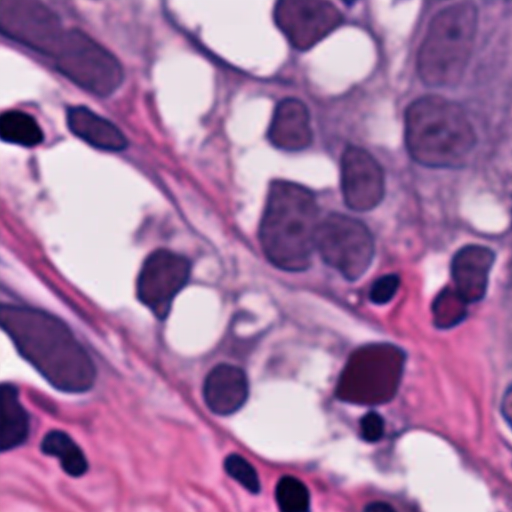}};
\node[picture format, anchor=north west] (A4) at (A3.north east)
    {\includegraphics[width=0.88in,height=0.88in]{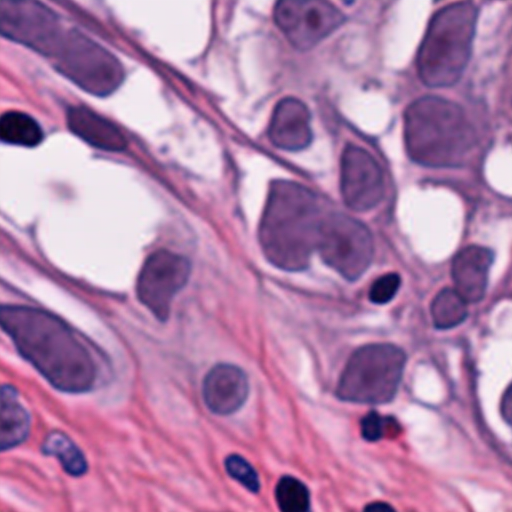}};
\node[picture format, anchor=north west] (A5) at (A4.north east)
    {\includegraphics[width=0.88in,height=0.88in]{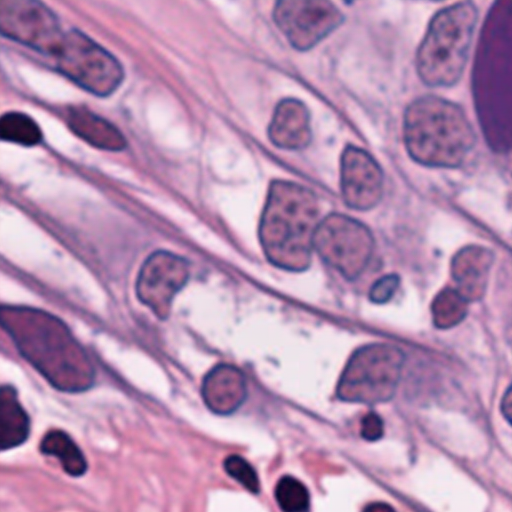}};
\node[picture format, anchor=north west] (A6) at (A5.north east)
    {\includegraphics[width=0.88in,height=0.88in]{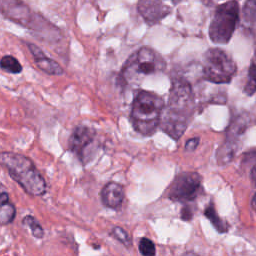}};

\node[picture format, anchor=north] (B1) at (A1.south)
    {\includegraphics[width=0.88in,height=0.88in]{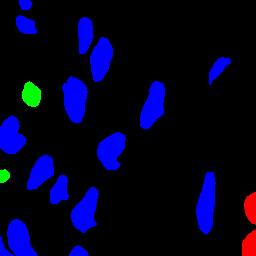}};
\node[picture format, anchor=north] (B2) at (A2.south)
    {\includegraphics[width=0.88in,height=0.88in]{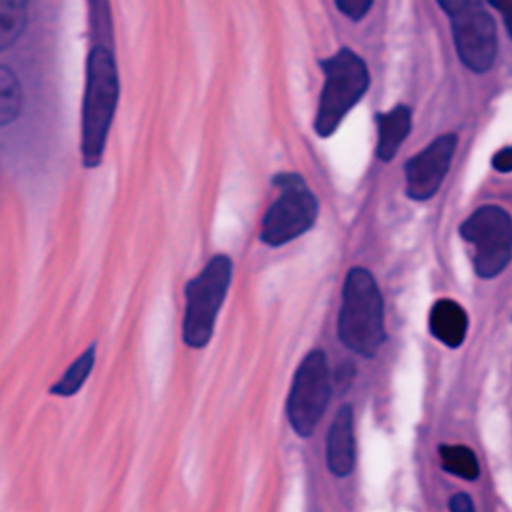}};
\node[picture format, anchor=north] (B3) at (A3.south)
    {\includegraphics[width=0.88in,height=0.88in]{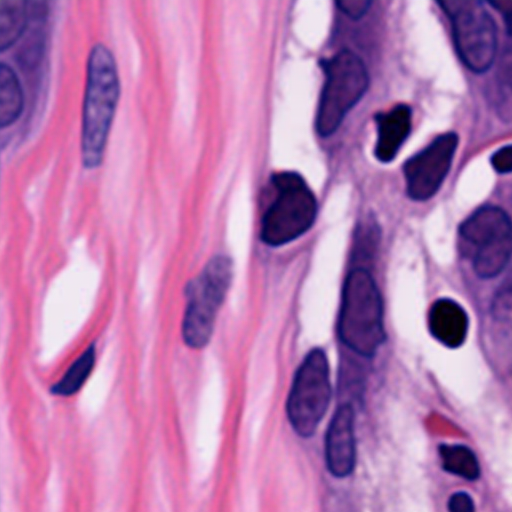}};
\node[picture format, anchor=north] (B4) at (A4.south)
    {\includegraphics[width=0.88in,height=0.88in]{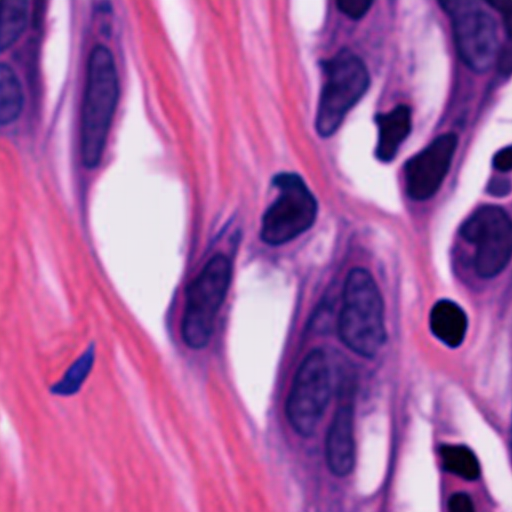}};
\node[picture format, anchor=north] (B5) at (A5.south)
    {\includegraphics[width=0.88in,height=0.88in]{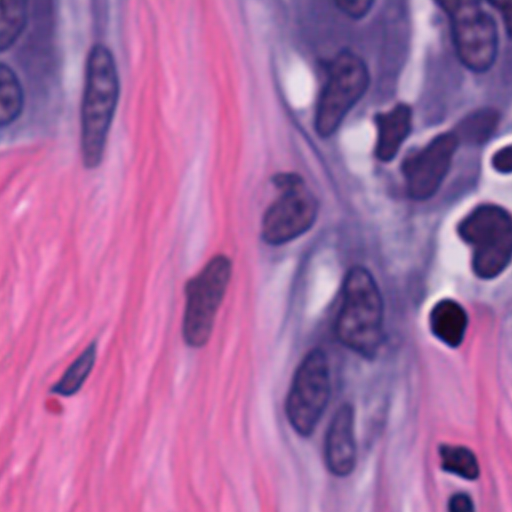}};
\node[picture format, anchor=north] (B6) at (A6.south)
    {\includegraphics[width=0.88in,height=0.88in]{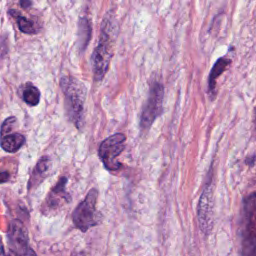}};

\node[picture format, anchor=north] (C1) at (B1.south)
    {\includegraphics[width=0.88in,height=0.88in]{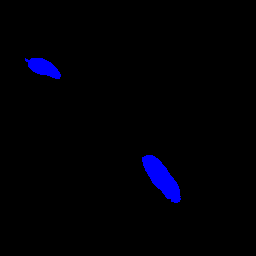}};
\node[picture format, anchor=north] (C2) at (B2.south)
    {\includegraphics[width=0.88in,height=0.88in]{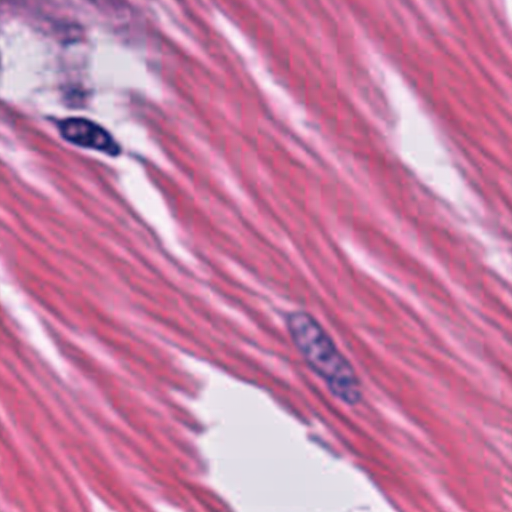}};
\node[picture format, anchor=north] (C3) at (B3.south)
    {\includegraphics[width=0.88in,height=0.88in]{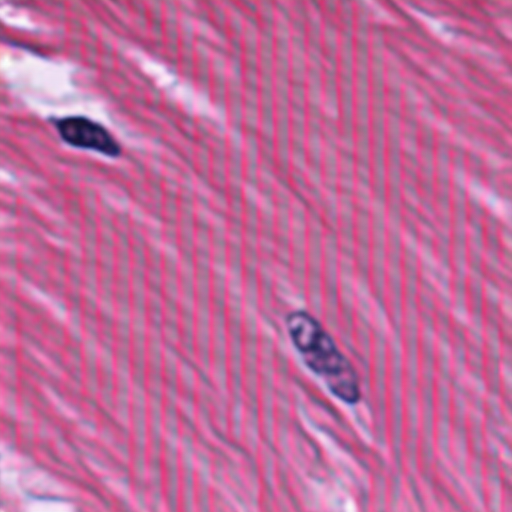}};
\node[picture format, anchor=north] (C4) at (B4.south)
    {\includegraphics[width=0.88in,height=0.88in]{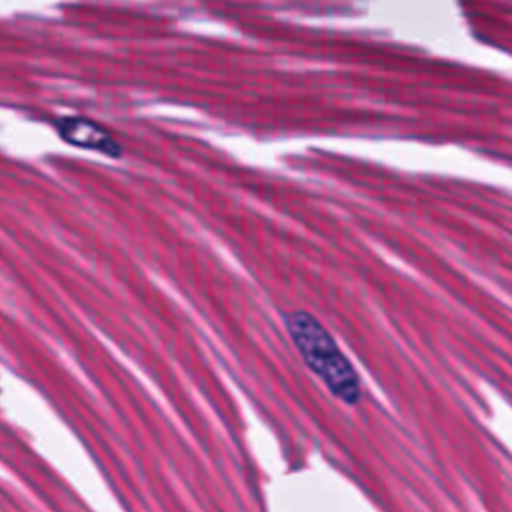}};
\node[picture format, anchor=north] (C5) at (B5.south)
    {\includegraphics[width=0.88in,height=0.88in]{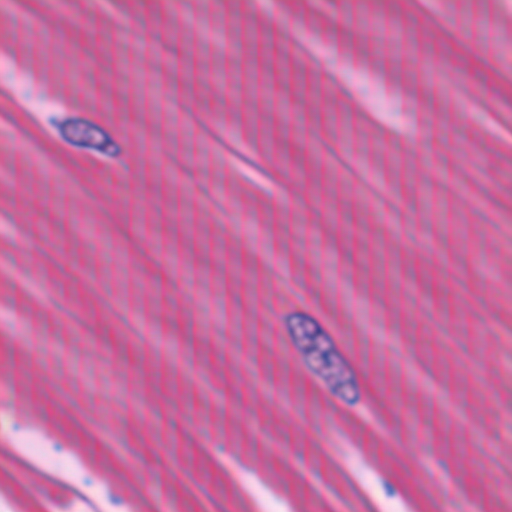}};
\node[picture format, anchor=north] (C6) at (B6.south)
    {\includegraphics[width=0.88in,height=0.88in]{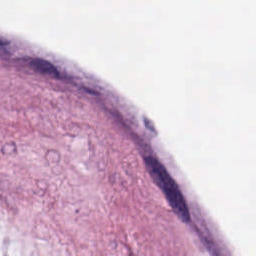}};

\node[anchor=south] at (A1.north) {\scriptsize \textbf{Mask}};
\node[anchor=south] at (A2.north) {\scriptsize \textbf{CN}};
\node[anchor=south] at (A3.north) {\scriptsize \textbf{SUM}};
\node[anchor=south] at (A4.north) {\scriptsize \textbf{MaxMin}};
\node[anchor=south] at (A5.north) {\scriptsize \textbf{CRM}};
\node[anchor=south] at (A6.north) {\scriptsize \textbf{Original}};


\end{tikzpicture}
\caption{Additional PanNuke qualitative comparisons. All generated columns use the same conditioning mask
as the corresponding original target image.}
\label{fig:qualitative_pannuke_supp}
\end{figure}

\begin{figure}[h]
    \centering
    \setlength{\tabcolsep}{1.5pt}
    \begin{tabular}{@{}cccccc@{}}
        & \scriptsize Input Mask & \scriptsize ControlNet & \scriptsize ORM & \scriptsize CRM (ours) & \scriptsize Original \\
        \multirow{3}{*}{\rotatebox[origin=c]{90}{\scriptsize}} &
        \includegraphics[width=0.18\textwidth]{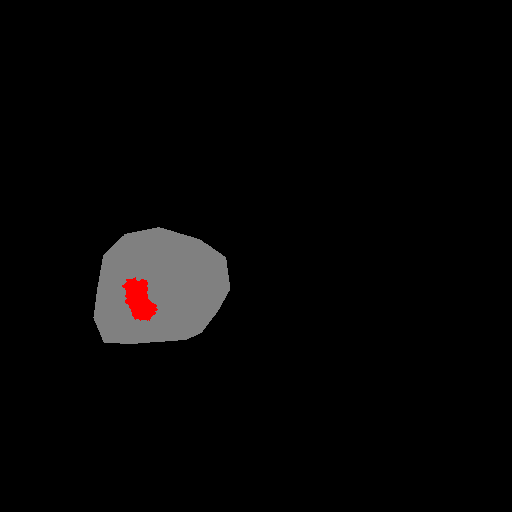} &
        \includegraphics[width=0.18\textwidth]{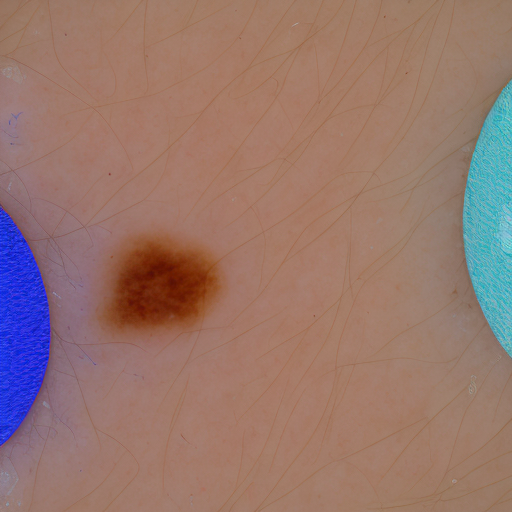} &
        \includegraphics[width=0.18\textwidth]{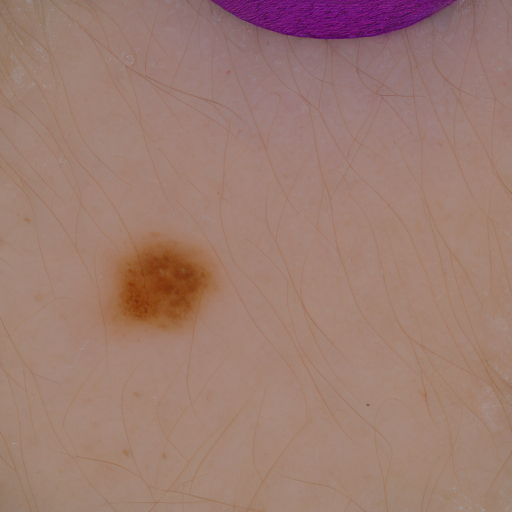} &
        \includegraphics[width=0.18\textwidth]{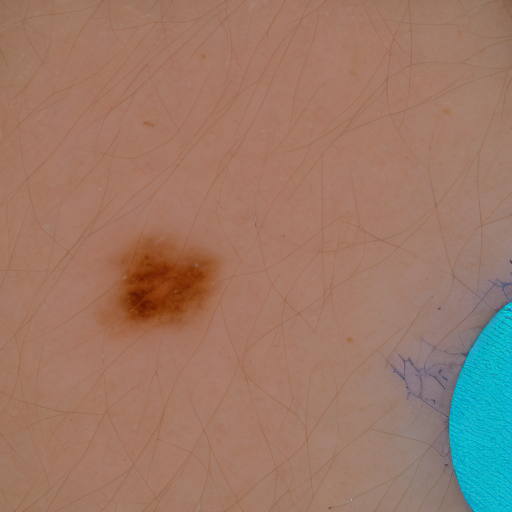} &
        \includegraphics[width=0.18\textwidth]{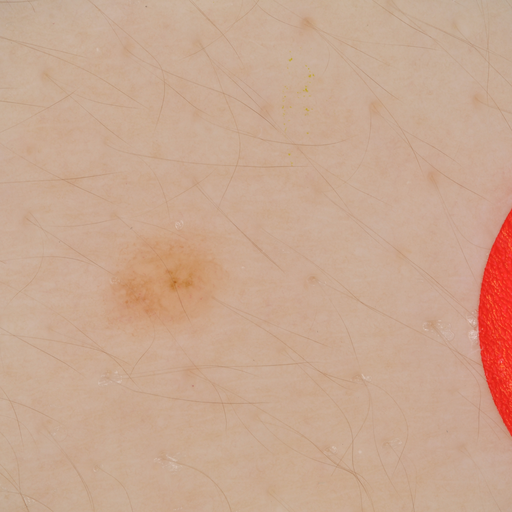} \\
        &
        \includegraphics[width=0.18\textwidth]{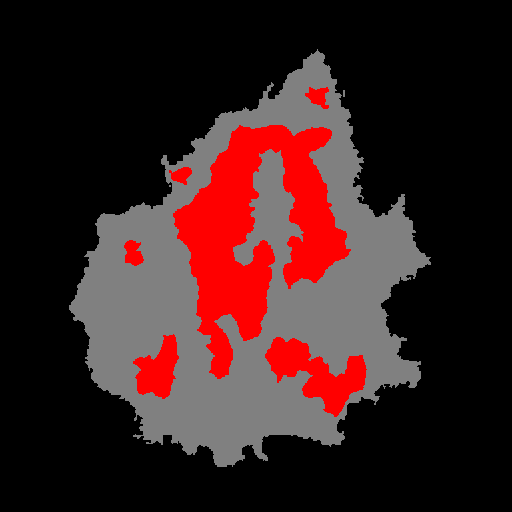} &
        \includegraphics[width=0.18\textwidth]{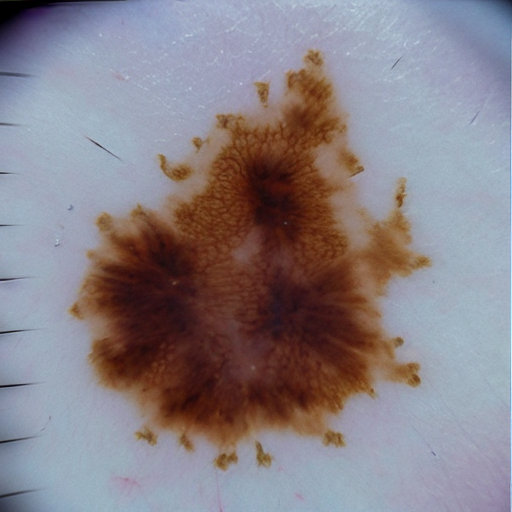} &
        \includegraphics[width=0.18\textwidth]{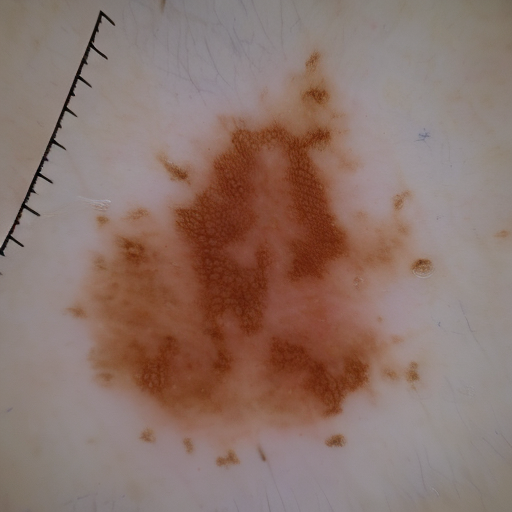} &
        \includegraphics[width=0.18\textwidth]{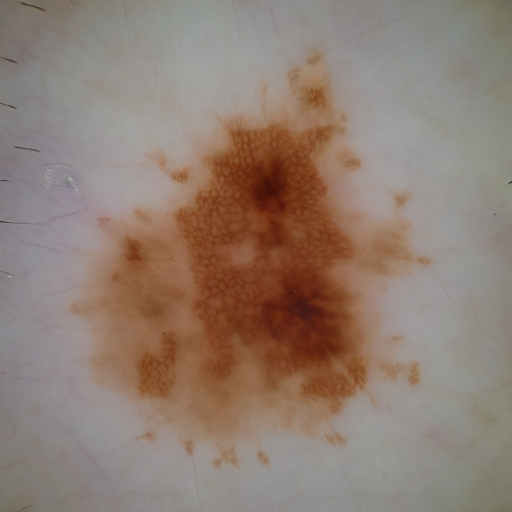} &
        \includegraphics[width=0.18\textwidth]{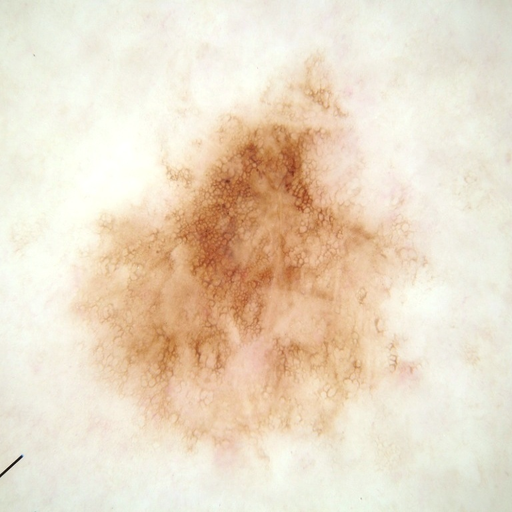} \\
        &
        \includegraphics[width=0.18\textwidth]{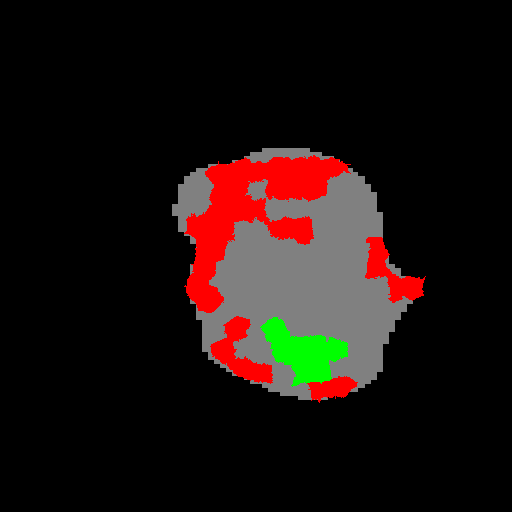} &
        \includegraphics[width=0.18\textwidth]{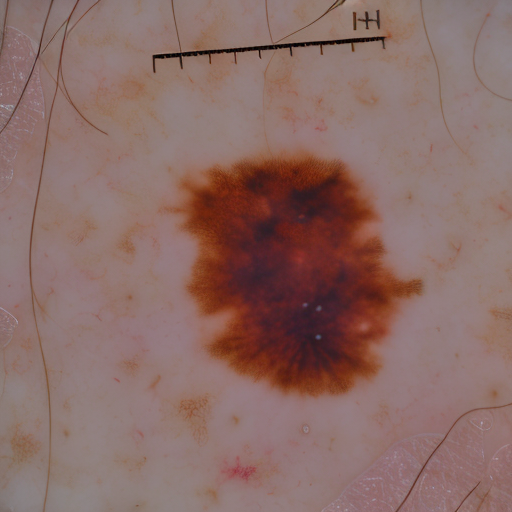} &
        \includegraphics[width=0.18\textwidth]{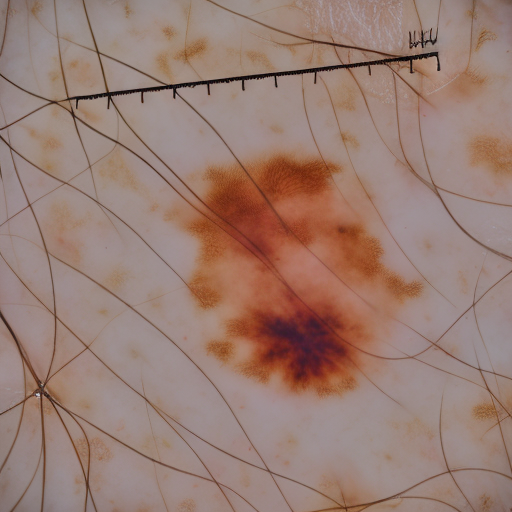} &
        \includegraphics[width=0.18\textwidth]{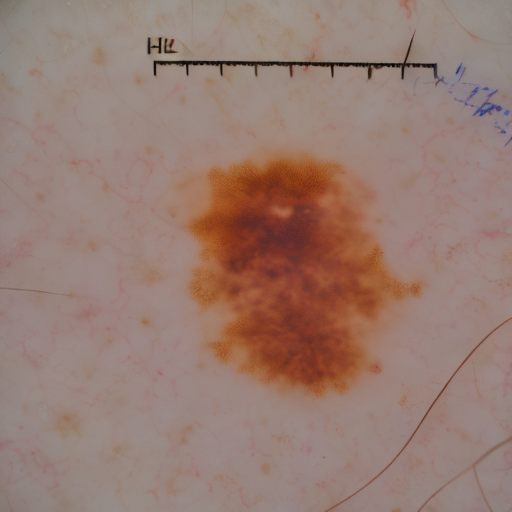} &
        \includegraphics[width=0.18\textwidth]{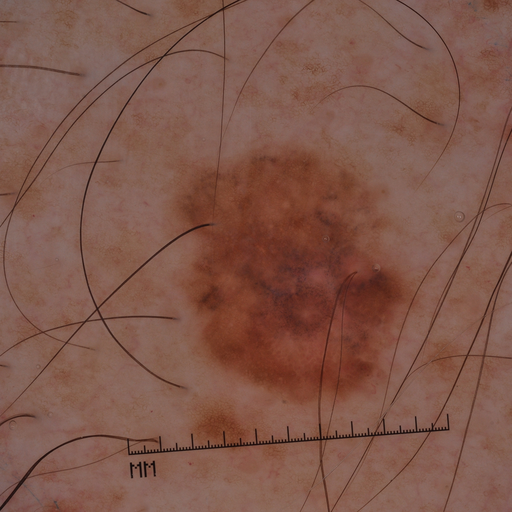}
    \end{tabular}
    \caption{Additional qualitative ISIC comparisons. Each row
    shows the input mask, generations from the SFT ControlNet,
    ORM-aligned model, and CRM-aligned model, and the original
    image under the same conditioning.}
    \label{fig:qualitative_isic_supp}
\end{figure}

Figures~\ref{fig:qualitative_pannuke_supp}, \ref{fig:qualitative_cedem_supp} and \ref{fig:qualitative_isic_supp} present additional qualitative comparisons for PanNuke, CeDeM, and ISIC.

\end{document}